\documentclass[lettersize,journal]{IEEEtran}
\usepackage{amsmath,amsfonts}
\usepackage{algorithmic}
\usepackage{algorithm}
\usepackage[algo2e]{algorithm2e}
\usepackage{array}
\usepackage[caption=false,font=normalsize,labelfont=sf,textfont=sf]{subfig}
\usepackage{textcomp}
\usepackage{stfloats}
\usepackage{url}
\usepackage{verbatim}
\usepackage{multirow}
\usepackage{graphicx}
\usepackage{cite}
\usepackage{booktabs}
\newtheorem{definition}{Definition}
\begin{document}

\title{Enhanced Urban Region Profiling with Adversarial Self-Supervised Learning for Robust Forecasting and Security}

\author{Weiliang Chen, Qianqian Ren, Yong Liu, Jianguo Sun
\thanks{Manuscript received April 19, 2021; revised August 16, 2021. This work was supported in part by the China Postdoctoral Science Foundation under Grant No.2022M711088. \textit{(Corresponding author: Qianqian Ren.)}

Weiliang Chen, Qianqian Ren and Yong Liu are with the Department of Computer Science and Technology, Heilongjiang University, Harbin, China. 
Jianguo Sun is with Hangzhou Institute of Technology, Xidian University, Hangzhou, China (e-mail:chanweiliang@s.hlju.edu.cn, renqianqian@hlju.edu.cn; liuyong123456@hlju.edu.cn, jgsun@xidian.edu.cn). 
}
}

\markboth{Journal of \LaTeX\ Class Files,~Vol.~14, No.~8, August~2021}%
{Shell \MakeLowercase{\textit{et al.}}: A Sample Article Using IEEEtran.cls for IEEE Journals}


\maketitle

\begin{abstract}
Urban region profiling plays a crucial role in forecasting and decision-making in the context of dynamic and noisy urban environments. Existing methods often struggle with issues such as noise, data incompleteness, and security vulnerabilities. This paper proposes a novel framework, Enhanced Urban Region Profiling with Adversarial Self-Supervised Learning (EUPAS), to address these challenges. By combining adversarial contrastive learning with both supervised and self-supervised objectives, EUPAS ensures robust performance across various forecasting tasks such as crime prediction, check-in prediction, and land use classification. To enhance model resilience against adversarial attacks and noisy data, we incorporate several key components, including perturbation augmentation, trickster generator, and deviation copy generator. These innovations effectively improve the robustness of the embeddings, making EUPAS capable of handling the complexities and noise inherent in urban data. Experimental results show that EUPAS significantly outperforms state-of-the-art methods across multiple tasks, achieving improvements in prediction accuracy of up to 10.8\%.
Notably, our model excels in adversarial attack tests, demonstrating its resilience in real-world, security-sensitive applications. This work makes a substantial contribution to the field of urban analytics by offering a more robust and secure approach to forecasting and profiling urban regions. It addresses key challenges in secure, data-driven modeling, providing a stronger foundation for future urban analytics and decision-making applications.
\end{abstract}


\section{Introduction}
Urban region profiling is crucial for understanding and managing various urban dynamics, including traffic flow, energy consumption, and public safety. With the proliferation of sensor networks and the increasing volume of urban data, accurate prediction models are essential for effective urban planning and decision-making. However, the vast and heterogeneous nature of urban data, combined with the complexity of urban systems, presents significant challenges for traditional modeling approaches. Additionally, urban systems are increasingly vulnerable to adversarial attacks, which can distort predictions and compromise the security and efficiency of urban management systems.

Urban region embeddings have been widely applied in diverse fields, including crime prediction \cite{Wang_2017_HDGE, Zhang_2020_MVGRE}, socio-demographic prediction \cite{Jean_2018_Tile2Vec}, and land usage classification \cite{Yao_2018_ZeMob}. These methods leverage diverse urban data sources—such as mobility data, Points of Interest (POIs), and socio-demographic information—to construct embeddings that capture the underlying characteristics of urban regions. However, most existing methods fail to account for the noise and incompleteness inherent in urban data, leading to suboptimal embeddings and, consequently, less reliable predictions.

\begin{figure*}[!t]
\centering
\subfloat[]{\includegraphics[width=3.3in]{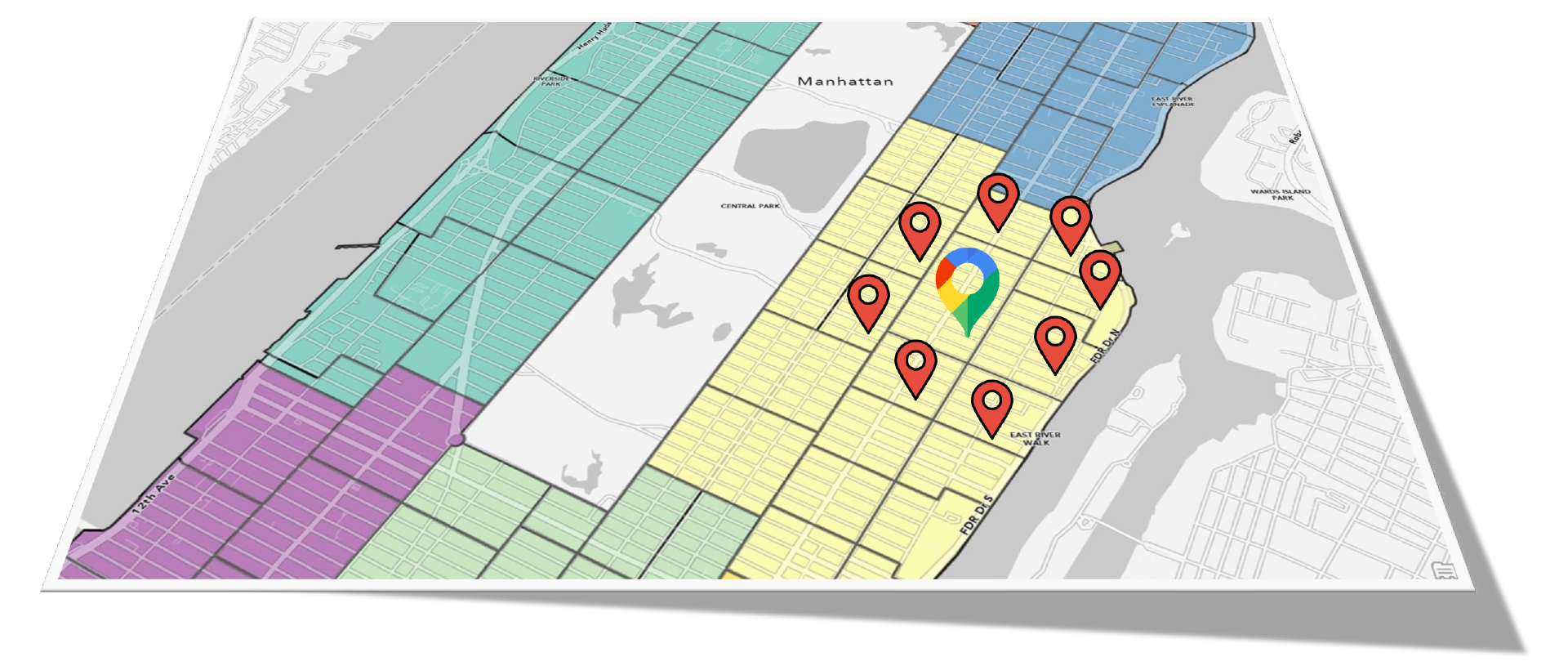}%
\label{IRnei}}
\hfil
\subfloat[]{\includegraphics[width=3.3in]{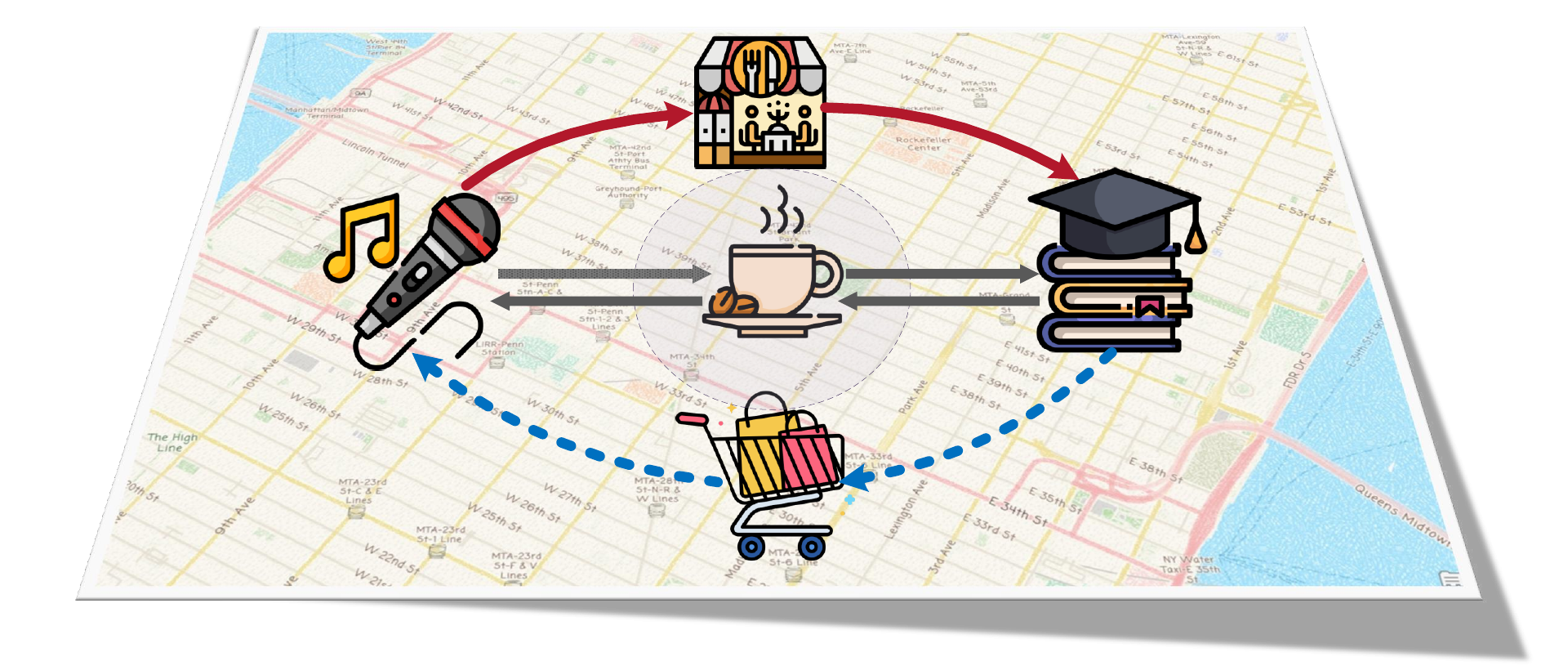}%
\label{AuMob1}}
\caption{Difficulties in Data Augmentation for Urban Data: (a) Distribution of a POI and its neighboring locations; (b) Replacing a POI (e.g., a café) with a nearby location (e.g., a restaurant or shopping center) fundamentally changes the semantics and purpose of the trip, distorting the model’s learning process.}\label{AuMob}
\label{intro}
\end{figure*}

Contrastive learning, a form of self-supervised learning, has shown great promise in learning robust representations by focusing on high-level semantics while ignoring irrelevant noise. It has achieved notable success in natural language processing \cite{GPT, GPT2} and computer vision \cite{MoCO, SimCLR}, with recent applications extending to graph-based data \cite{TKDE_SSL_sur}. However, applying contrastive learning to urban region profiling faces several challenges due to the complexity and heterogeneity of urban data. One major challenge lies in data augmentation, a key component of contrastive learning. Existing augmentation methods, such as masking or cropping, often fail to generate semantically meaningful positive samples in the urban context. Instead, these methods risk distorting the semantics of points of interest (POIs) and mobility patterns. For instance, as illustrated in Fig.~\ref{AuMob}, replacing a POI (e.g., a café) with a nearby location (e.g., a restaurant or shopping center) fundamentally changes the semantics and purpose of the trip. Such distortions can significantly hinder the model's ability to learn robust representations, highlighting the inadequacy of traditional augmentation strategies in urban region profiling.

Building on these observations, it is evident that current urban region profiling models face several unresolved issues:
\begin{itemize}
 \item \textbf{Vulnerability to Adversarial Attacks:} Existing models  prioritize accuracy and efficiency but neglect security considerations. This oversight leaves them ill-prepared to handle adversarial attacks or malicious distrubances, which can undermine the reliability of smart city applications.
 \item \textbf{Lack of Robustness:} Traditional data augmentation methods, such as substitution and cropping, often fail to generate effective and semantically consistent samples. This limitation becomes particularly problematic in dynamic, uncertain, and noisy urban environments, resulting in suboptimal model performance in complex real-world scenarios.
 \item \textbf{Limited Integration of Multi-source Data:} Many models rely on a single data source, such as human mobility data or POIs, limiting the understanding of urban systems and failing to exploit the rich interrelations among diverse data sources.
\end{itemize}

To address these challenges, we integrate adversarial training with contrastive learning into the urban region profiling framework. Adversarial training introduces tailored perturbations into the learning process, simulating adversarial scenarios and enhancing the robustness of the model. These perturbations act as challenging negative samples within the contrastive learning framework, while carefully generated augmentations serve as positive samples. This combination ensures the model learns robust and discriminative embeddings that capture high-level semantic relationships and maintain resilience against adversarial attacks. By leveraging this integrated approach, we significantly improve urban region profiling tasks such as check-in prediction, land usage classification, and crime prediction, even in the presence of noisy or incomplete data.

In this paper, we propose EUPAS, a robust and secure framework for urban region profiling that combines adversarial contrastive learning and supervised learning. Our approach addresses critical challenges, including data noise, adversarial vulnerability, and insufficient data augmentation, while enabling the integration of multi-source urban data. We validate the effectiveness of our framework through comprehensive experiments on tasks such as check-in prediction, land usage classification, and crime prediction. The experimental results show that EUPAS not only achieves superior prediction accuracy but also enhances resilience to adversarial attacks and provides improved security for urban management systems.

To sum up, our work has the following contributions:
\begin{itemize}
\item We introduce a novel framework for urban region profiling that integrates adversarial self-supervised learning with perturbation augmentation. This combination improves the robustness of urban forecasting models by enhancing the quality of urban region embeddings, ensuring better performance in dynamic, noisy, and incomplete urban environments.

\item We employ an adversarial strategy to automatically generate challenging positive-negative pairs for contrastive learning. By incorporating advanced adversarial defense mechanisms, our approach enhances the model’s ability to capture high-level semantics and secures urban region embeddings against potential adversarial manipulations. This results in a safer and more reliable system for urban management.

\item  Our model outperforms state-of-the-art baselines in key urban forecasting tasks, including traffic prediction, land usage classification, and crime prediction. It significantly enhances forecasting accuracy and addresses challenges such as data noise and incompleteness, making it highly effective for real-world urban analytics.

\item Through extensive experiments on real-world data, we demonstrate that our model not only surpasses existing models in performance but also successfully withstands adversarial attack tests. This highlights its robustness and reliability, particularly in handling adversarial distrubances and the inherent uncertainties of urban data.
\end{itemize}

\section{Related Work}

\subsection{Region Representation Learning}

The growing availability of diverse urban data sources has spurred significant research on learning urban region representations. Various approaches have been proposed to enhance the effectiveness of region embeddings for urban data analysis.
MV-PN \cite{Fu_2019_MVPN} introduced a region embedding model that captures both intra- and inter-regional similarities using point-of-interest (POI) networks and spatial autocorrelation layers. Building on this, CGAL \cite{CGAL} extended the approach by introducing a collective adversarial training strategy, further improving the quality of region representations. MVGRE \cite{Zhang_2020_MVGRE} developed a multi-view joint learning model to learn region embeddings, capturing diverse information from multiple perspectives. MGFN \cite{Wu_2022_MGFN} focused on extracting traffic patterns for region embeddings, but its model is limited to mobility data and does not incorporate POI data, which is crucial for understanding the functional characteristics of urban regions.
ROMER \cite{ROMER} improved urban region embedding by capturing multi-view dependencies from diverse data sources, leveraging global graph attention networks, and employing a two-stage fusion module to combine the learned features. HREP \cite{HREP} applied a continuous prompt method, prefix-tuning, to replace the direct use of region embeddings in downstream tasks. While these models have achieved significant advancements, they often fail to address the vulnerability of embeddings to adversarial disturbances. Noisy or malicious data can severely degrade the performance of these models, as they lack mechanisms to protect the embedding process from adversarial manipulations.

In response to these challenges, recent research has explored integrating adversarial robustness into region representation learning. Techniques such as adversarial augmentation \cite{advaugmenre} and robust contrastive training \cite{robustCL} have shown promise in improving the resilience of learned embeddings. However, their application to urban region profiling is still limited, especially considering the complex and dynamic nature of urban data.

\subsection{Graph Neural Network for Region Representation}

Graph Neural Networks (GNNs) have become a prominent tool for learning graph embeddings, particularly in the context of complex graph structures. The rapid development of GNNs \cite{Kipf_2016_Variational, Kipf_2017_SemiSupervised} has been accompanied by a growing interest in improving their robustness to adversarial attacks. Notably, models like RGCN \cite{R-GCNs} and HetGNN \cite{HetGNN} extend GNNs to heterogeneous graphs, enabling them to handle more complex data types. However, these models typically assume clean graph data during training, making them vulnerable to adversarial manipulations.

Adversarial attacks on graph structures commonly involve perturbing node attributes, graph structures, or edge connections \cite{ADGSTUDY}. Defense strategies, such as those proposed by AD-GCL \cite{ADGCL} and RGCN \cite{RGCN}, aim to mitigate these attacks, but they are largely tailored to homogeneous graphs and do not address the specific challenges posed by complex, real-world urban data. In addition, while methods like GraphSage \cite{Hamilton_2018_GraphSAGE} and GAT \cite{Velickovic_2018_GAT} are successful in inductive learning and incorporating attention mechanisms, they do not explicitly focus on adversarial robustness.

In the context of urban region profiling, the security of embeddings against adversarial attacks remains an open challenge. Existing methods often fail to integrate adversarial defenses with the inherent characteristics of graph-based urban data, thus compromising the reliability and security of embeddings.

\subsection{Graph Contrastive Learning for Region Representation}

Contrastive learning has proven effective in fields such as natural language processing \cite{CILP} and computer vision \cite{MoCO, BYOL}, and its application to graph data for self-supervised learning has gained attention. This approach excels at capturing high-level semantics by distinguishing between similar and dissimilar data points, making it an effective tool for learning robust embeddings. However, contrastive learning methods are particularly susceptible to adversarial attacks \cite{ADGSTUDY, Xu__InfoGCL}, as small, adversarial perturbations can undermine the model’s ability to generate meaningful representations.

Several techniques, such as edge perturbation \cite{GraphCL} and adversarial graph augmentation \cite{ADGCL}, have been proposed to address this vulnerability, but they are often tested on synthetic or controlled datasets that do not fully reflect the complexities of real-world urban environments. Recent work, including methods like InfoGCL \cite{Xu__InfoGCL} and AD-GCL \cite{ADGCL}, has explored adversarial contrastive learning to enhance model robustness by generating hard negatives through adversarial perturbations. These approaches help improve model discrimination by encouraging the model to focus on harder-to-distinguish examples. However, their application to urban region profiling remains limited, as urban graphs involve heterogeneous data sources and dynamic interactions that require more tailored solutions.

Motivated by these findings, we propose leveraging adversarial contrastive learning specifically designed for urban region data. By incorporating adversarial perturbations into the contrastive transformation process, our approach aims to generate robust and secure embeddings that are resilient to adversarial threats. This adaptation is particularly important in the context of urban region profiling, where the data is often noisy, incomplete, and subject to security vulnerabilities.
\section{Preliminaries}

In this section, we introduce key notations and define the urban region embedding problem. Consider a city partitioned into \( N \) non-overlapping regions. We define a graph \( \mathcal{G} = (\mathcal{V}, \mathcal{E}) \), where \( \mathcal{V} \) denotes the set of region nodes, with each node corresponding to a specific urban region, and \( \mathcal{E} \) is the set of edges connecting the nodes. We assume that there are \( K \) distinct relations (where \( K > 1 \)) between two nodes. The edges corresponding to the \( k \)-th relation are denoted by \( \mathcal{E}_k \), where \( 1 < k \leq K \). In the following sections, we exploit \( \mathcal{E}_k \) for each relation to generate a subgraph \( \mathcal{G}_k = (\mathcal{V}, \mathcal{E}_k) \). These relations are based on human mobility data correlations, Points of Interest (POI) information, and the geographic neighbors of each region.

\begin{definition}[\bf Human Mobility]
Human mobility is characterized as a collection of trips within the urban regions. Let \( t = (r_o, r_t) \) represent a travel record, where \( r_o \) is the origin region and \( r_t \) is the destination region, with \( 1 \leq o,t \leq K \). The set of all trips is denoted as \( \mathcal{T} = \{\overrightarrow{t_0}, \overrightarrow{t_1}, \ldots, \overrightarrow{t_{|M|}}\} \), where \( M \) is the total number of trips.
\end{definition}

\begin{definition}[\bf POI Information]
The functional characteristics of each region are represented by its Points of Interest (POIs). We denote the POI information for all regions as \( \mathcal{P} = \{p_1, p_2, \ldots, p_K\} \in \mathbb{R}^{f \times K} \), where \( f \) is the number of POI categories, and each \( p_i \) represents the POI information of region \( r_i \).
\end{definition}

\begin{definition}[\bf Geographic Neighbors]
Geographic neighbor information encodes the spatial relationships between a region and its neighboring regions. We denote the set of geographic neighbors for each region as \( \mathcal{N} = \{\overrightarrow{n_1}, \overrightarrow{n_2}, \ldots, \overrightarrow{n_K}\} \), where \( \overrightarrow{n_i} = (r_1, r_2, \ldots, r_{|\overrightarrow{n_i}|}) \) represents the geographic neighbors for region \( r_i \).
\end{definition}

\textbf{Problem Definition:}  
Given the graph \( \mathcal{G} \), human mobility data \( \mathcal{T} \), POI information \( \mathcal{P} \), and geographic neighbor data \( \mathcal{N} \), our goal is to learn low-dimensional embeddings \( \mathcal{E} = \{e_1, e_2, \ldots, e_N\} \), where each \( e_i \in \mathbb{R}^d \) is the \( d \)-dimensional embedding of region \( r_i \in \mathcal{V} \). These embeddings should preserve critical information from mobility patterns, geographic relationships, and POI features, which are essential for downstream urban tasks such as traffic forecasting, land usage classification, and crime prediction.

\begin{figure*}
\setlength{\abovecaptionskip}{-0.0080cm}
\setlength{\belowcaptionskip}{-0.3cm}
\centering\includegraphics[width=0.95\linewidth]{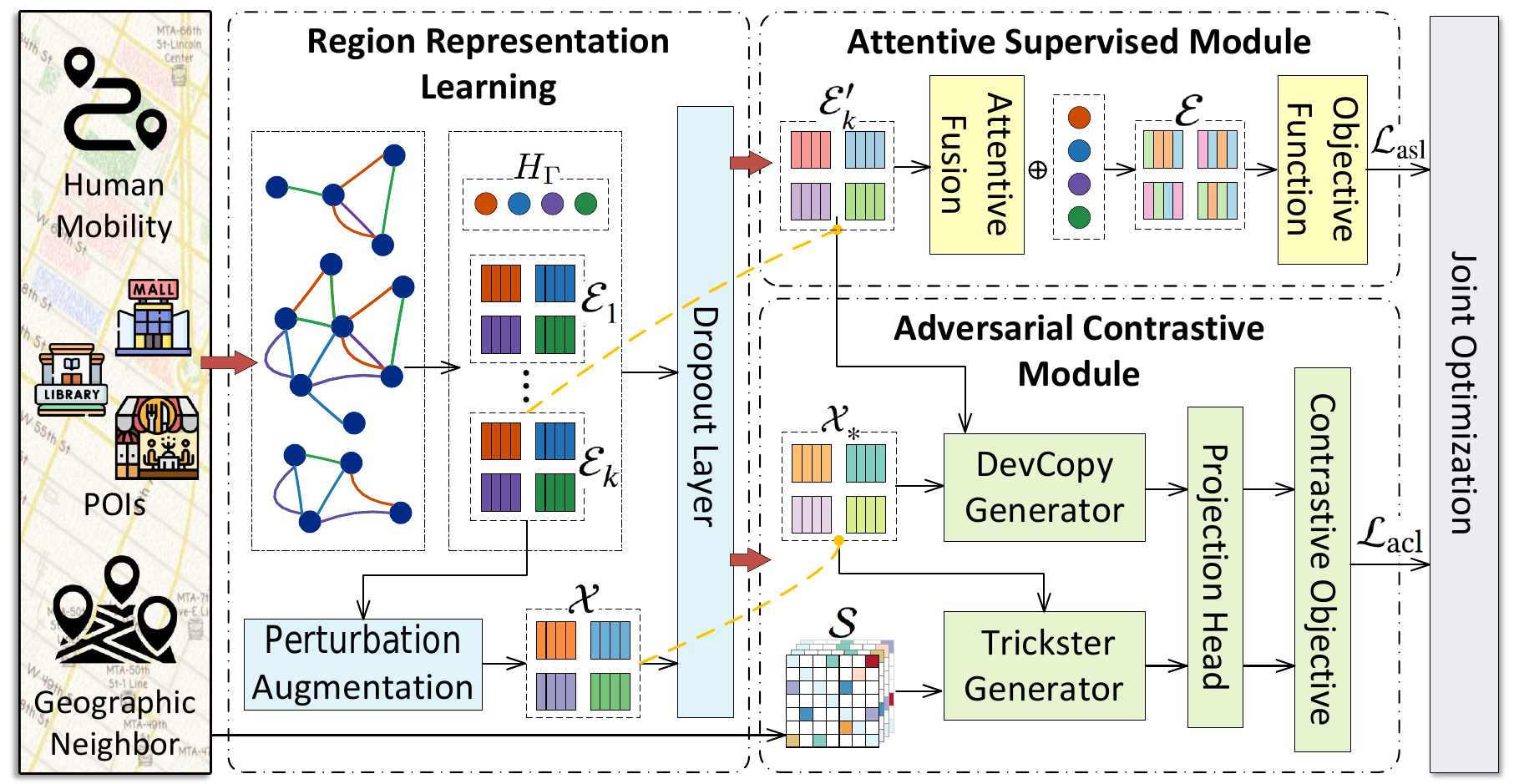}
    \caption{The architecture of EUPAS integrates four key components: Region Representation Learning, Perturbation Augmentation, Attentive Supervised Module, and Adversarial Contrastive Learning. }
   \label{architecture}
\end{figure*}

\begin{table}
\caption{List of important notations.}
\centering
\label{tab:not}
\scalebox{1}{\begin{tabular}{@{}c|l@{}}
\toprule
$N$            & the number of regions                \\
$K$            & the number of relation types                \\
$\mathcal{S}$         & similarity matrix for different data sources \\ \midrule
$H_\Gamma$              & relation embedding representing inter-region dependencies                   \\
$\mathcal{E}_k$  & relation-focused region embedding for region $k$   \\
$\mathcal{E}_k^{\prime}$  & anchor embedding for region $k$                    \\
$\mathcal{E}$ & final task-specific region embedding                     \\ \midrule
$\mathcal{X}_{*}$           & perturbation-augmented node embedding (DA-NE)                   \\
${\mathcal{X}}^-_{*}$     & hard negative samples (generated by Trickster)        \\
${\mathcal{X}}^+_{*}$     & strong positive samples (generated by DevCopy)        \\ \midrule
$\mathcal{L}_{\text{acl}}$ & adversarial contrastive loss function                \\
$\mathcal{L}_{\text{asl}}$  & attentive supervised loss function                    \\
$\mathcal{L}_{\text{total}}$  & total loss function                    \\
$\tau$ & temperature coefficient for contrastive loss scaling  \\
$\alpha$ & balancing factor for positive and negative samples in contrastive loss \\
$\beta$ & regularization factor for total loss \\
$\gamma$ & adversarial perturbation coefficient  \\ \bottomrule
\end{tabular}}
\end{table}

\section{Methodology}
\subsection{Framework Overview}
The overall model architecture is depicted in Fig.~\ref{architecture}. This section presents our proposed approach for robust urban region profiling, which integrates four key components: Region Representation Learning, Perturbation Augmentation, Attentive Supervised Module, and Adversarial Contrastive Learning. These components work synergistically to enhance the model’s robustness, generalization, and security, enabling it to effectively handle the challenges posed by dynamic, noisy, and uncertain urban environments.

The Region Representation Learning module extracts high-quality embeddings that capture the spatial and temporal dynamics of urban regions, forming the foundation for all downstream tasks. The Perturbation Augmentation module then introduces controlled noise into these embeddings, simulating real-world urban noise, and enhancing the model’s resilience against overfitting and adversarial perturbations.

Next, the Attentive Supervised Module refines these embeddings through attention mechanisms, prioritizing important features while filtering out irrelevant noise. Finally, the Adversarial Contrastive Learning module strengthens the model’s ability to differentiate between semantically similar and dissimilar data by generating hard positive-negative pairs via adversarial perturbations.

Collectively, these components form a robust framework that not only improves predictive accuracy in urban forecasting tasks but also ensures the model is secure against adversarial attacks. In the following subsections, we provide detailed descriptions of each component and explain their interactions, highlighting how they contribute to enhancing the overall performance of the model. The important notations are listed in Table.~\ref{tab:not}.

\subsection{Region Representation Learning}
Firstly, we learn the embedding representations of regions under each relation. Let $\mathcal{E}_k^{(l)}=\left\{ e_{1k}^{(l)}, e_{2k}^{(l)}, \ldots, e_{Nk}^{(l)}\right\}$ denote the region embedding under $k$-th relation, $H_\Gamma^{(l)}=\left\{h_0^{(l)}, h_1^{(l)}, \cdots, h_K^{(l)}\right\}$ denote relation embeddings at the $l$-th layer, respectively.
Then, given the initial node embedding $\mathcal{E}_k^{(0)}$ and relation embedding $H_\Gamma^{(0)}$, we update them at the $l-${th} layer as follows:
\begin{equation}
e_{uk}^{(l)}=\sigma\left(\sum_{\gamma \in \Gamma} \sum_{v \in \mathcal{N}_u^\gamma} \phi_{u, \gamma}  \mathbf{W}^{(l)}( e_{vk}^{(l-1)}\circ h_k^{(l-1)})\right),
\end{equation}
\begin{equation}
 h_k^{(l)}=\mathbf{W}_k^{(l)} h_k^{(l-1)}+\mathbf{b}_k^{(l)},
\end{equation}
where $u,v \in \{1,2,\cdots,N\}$, $k \in \{0,1,\cdots,K\}$.
$\sigma(\cdot)$ represents the LeakyReLU activation function, and $\circ$ is the element-wise product. $\mathbf{W}^{(l)}\in \mathbb{R}^{d \times d}$ is a learnable parameter, $\mathcal{N}_u^k$ stands for the neighbors set of region $v_u$ under $k$-th relation. $\mathbf{W}_k^{(l)}$ and $\mathbf{b}_k^{(l)}$ are layer-specific parameters projecting relation embeddings from the previous layer to the same embedding space for subsequent use.
$\phi_{u, k}=\frac{1}{\left|\mathcal{N}_u^k\right|}$ is the normalization weight, which quantifies the correlation between region pair $\left(v_i, v_j\right)$.

To mitigate feature smoothing in graph neural networks, we incorporate ResNet\cite{ResNet} for multi-layer graph aggregation. At the $l-$th layer, the aggregation process is expressed as follows:
\begin{equation}
\mathcal{E}_k^{(l+1)}=\mathcal{E}_k^{(l)}+\sigma\left({D}^{-\frac{1}{2}} {A} {D}^{-\frac{1}{2}} \mathcal{E}_k^{(l)} \mathbf{W}^{(l)}\right),
\end{equation}
where ${A} \in \mathbb{R}^{|N| \times N|}$ denotes the adjacency matrix, and ${D} \in \mathbb{R}^{|N| \times N|}$ is the corresponding diagonal matrix.

To efficiently handle large-scale urban data, we store the adjacency matrices $\mathbf{A}_k$ (representing relations such as human mobility, POI data, and geographic neighbors) as sparse matrices. Sparse matrices only retain non-zero elements, significantly reducing memory consumption. For each region, we keep only the top $k$-nearest neighbors based on similarity measures for human mobility, POIs, and geographic neighbors, ensuring that most matrix elements are zero.

After aggregating information across different relations and applying spectral convolutions, we obtain relation-focused region embeddings $\mathcal{E}_k = \{\mathcal{E}_k^o, \mathcal{E}_k^t, \mathcal{E}_k^p, \mathcal{E}_k^g\}$ and corresponding relation embeddings $H_\Gamma = \{h_o, h_t, h_p, h_g\}$. 
These embeddings are then used for further learning and prediction tasks.

\subsection{Perturbation Augmentation for Robustness and Security}
In real-world applications, urban region data is often affected by noise, which can distort the underlying information, reducing the model’s reliability. This challenge is further compounded by data sparsity, which can lead to overfitting and create security vulnerabilities in predictive models. To address these issues, we propose a perturbation augmentation technique aimed at improving the robustness of the model against both noise and adversarial attacks by augmenting the graph data. This approach helps in boosting the model's generalization ability and safeguarding against noise-induced vulnerabilities.

The perturbation augmentation process involves perturbing the region embeddings, $\mathcal{E}_k$, at the $k$-th relation. Specifically, we apply a random perturbation to the embeddings as follows:

\begin{equation} \mathcal{X}_k = f\left(\mathcal{E}_k + \eta \Delta \boldsymbol{\theta}_{sp}\right), \quad \Delta \boldsymbol{\theta}_{sp} \sim \mathcal{N}(0, \boldsymbol{\sigma}), \end{equation}
where $\Delta \boldsymbol{\theta}_{sp}$ is a random perturbation sampled from a Gaussian distribution with zero mean and standard deviation $\sigma$, representing the noise added to the embeddings.
$\eta$ is a scaling factor that determines the intensity of the noise perturbation.
The function $f(\cdot)$ is a dropout layer, which applies stochastic masking to the embeddings to encourage generalization and improve robustness against overfitting and adversarial perturbations.

This augmented node embedding, $\mathcal{X}_k$, introduces noise into the model during training, improving its ability to handle data sparsity and noise in real-world scenarios. The perturbation augmentation process strengthens the model's resilience, ensuring it remains secure even in the presence of small, adversarial perturbations. Algorithm \ref{alg:perturbation_augmentation} further illustrates how perturbation augmentation works.

\begin{algorithm}[htbp]
\caption{Perturbation Augmentation for Robustness and Security}
\label{alg:perturbation_augmentation}
\KwIn{Region embedding $\mathcal{E}_k$, perturbation intensity $\eta$, standard deviation $\sigma$.}
\KwOut{Augmented embedding $\mathcal{X}_k$.}

\textbf{Stage 1: Sampling Perturbation} \\
Sample perturbation from Gaussian distribution:
\[
\Delta \boldsymbol{\theta}_{sp} \sim \mathcal{N}(0, \sigma)
\]

\textbf{Stage 2: Apply Scaling} \\
Scale the perturbation by the intensity factor:
\[
\Delta \boldsymbol{\theta}_{sp} \gets \eta \cdot \Delta \boldsymbol{\theta}_{sp}
\]

\textbf{Stage 3: Apply Dropout} \\
Generate dropout mask:
\[
DropoutMask \sim \text{Bernoulli}(p=0.5)
\]
Apply dropout to the scaled perturbation:
\[
\mathcal{X}_k \gets \mathcal{E}_k + DropoutMask \cdot \Delta \boldsymbol{\theta}_{sp}
\]

\KwRet{$\mathcal{X}_k$}
\end{algorithm}

\subsection{Attentive Supervised Module}\label{secASM}

To effectively incorporate the representation of regions under various relations, we propose an attentive supervised module that learns a weight coefficient for each relation. This coefficient dynamically captures the varying importance of different relations, enabling the model to focus on the most impactful factors. Furthermore, a supervised loss function guides the optimization process, enhancing the quality of the learned embeddings.

First, we define the semantic fusion coefficient $\alpha_k$ for the $k$-th relation as follows:
\begin{equation}
\alpha_k = \frac{1}{|N|} \sum_{j=1}^{|N|} \mathbf{q}^{\top} \cdot \sigma\left(\mathbf{W} {e}_{jk} + \mathbf{b}\right),
\end{equation}
where $\mathbf{q}$ is the attention vector, and $\sigma(\cdot)$ represents the LeakyReLU activation function. The parameters $\mathbf{q}$, $\mathbf{W}$, and $\mathbf{b}$ are shared across all region embeddings, ensuring a consistent projection into the same space for the computation of $\alpha_k$.

Then, using the learned coefficients $\alpha_k$, the final region representation integrates all relations as:
\begin{equation}
\mathcal{E} = \sum_{k=1}^K \operatorname{softmax}\left(\alpha_k\right) \cdot \mathcal{E}_k^{\prime} \cdot h_k,
\end{equation}
where $\mathcal{E}_k^{\prime}$ represents the region embeddings under the $k$-th relation, and $h_k$ is the corresponding relation embedding. This aggregation ensures that the representation is both comprehensive and focused on significant relationships.

Inspired by \cite{HREP}, we design a unified loss function to optimize the module, defined as:
\begin{align}
\mathcal{L}_{\text{asl}} = & \sum_{i=1}^{N}\max \left\{\left\|e_i - e_i^{+}\right\|_2 - \left\|e_i - e_i^{-}\right\|_2, 0\right\} \notag \\
& + \sum_{(r_i, r_j) \in \mathcal{M}} \left( \log \frac{\hat{d}_o(r_j \mid r_i)}{p_o(r_j \mid r_i)} + \log \frac{\hat{d}_t(r_i \mid r_j)}{p_t(r_i \mid r_j)} \right) \notag \\
& + \sum_{i=1}^{N} \sum_{j=1}^{N} \left[\mathcal{S}_p^{ij} - (e_{ip})^\top e_{jp}\right]^2,
\end{align}
where
 $e_i^{+}$ and $e_i^{-}$ are the positive and negative geographic neighbors of region $r_i$, respectively.
  $e_{ip}$ and $e_{jp}$ represent the embeddings of regions $r_i$ and $r_j$ under the POI relation.
    $\mathcal{S}_p$ denotes the POI similarity matrix, constructed following \cite{Zhang_2020_MVGRE}.
    $p_o(r_j \mid r_i)$ and $p_t(r_i \mid r_j)$ are the original mobility distributions.
    $\hat{d}_o(r_j \mid r_i)$ and $\hat{d}_t(r_i \mid r_j)$ are the reconstructed origin and target distributions, computed as:
   \begin{equation}
       \begin{split}
    \hat{d}_o(r_j \mid r_i) = \frac{\exp({e_i^{s}}^\top e_j^t)}{\sum_j \exp({e_i^{s}}^\top e_j^t)}, \\
    \hat{d}_t(r_i \mid r_j) = \frac{\exp({e_j^{t}}^\top e_i^s)}{\sum_i \exp({e_i^{t}}^\top e_i^s)}.
     \end{split}
 \end{equation}
This loss function balances the supervised learning objectives across multiple relationships, ensuring the embeddings capture the most relevant features while maintaining consistency with the observed data distributions.

\begin{figure*}[!t]
\centering\includegraphics[width=1.0\textwidth]{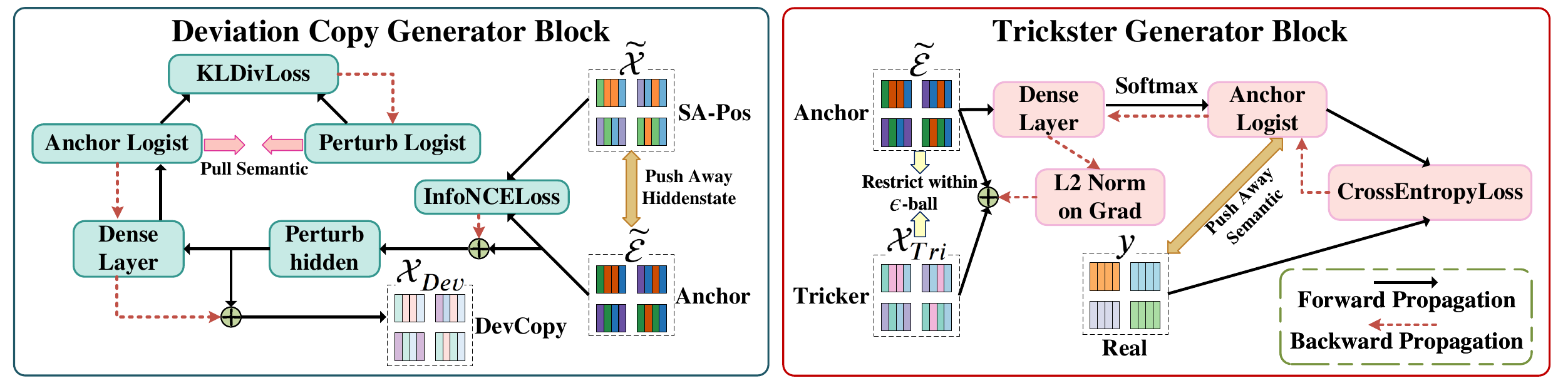}
\caption{Workflow diagram for the deviation copy and trickster generators.}
\label{Adv}
\end{figure*}

\subsection{Adversarial Contrastive Module}
To enhance the robustness and generalization of region embeddings, we propose an adversarial contrastive module, which utilizes adversarial perturbations to generate hard positive and negative pairs. This approach is designed to effectively address semantic biases and noise in region embeddings, aligning with the objectives of secure and reliable representation learning in urban data analysis.

\subsubsection{Adversarial Perturbation Layer}
Region embeddings often exhibit high intrinsic separability, which can impede the effectiveness of contrastive learning. To address this, we design an Adversarial Perturbation Layer that generates challenging positive and negative samples, exposing the model to harder training scenarios. The layer comprises two components: the Deviation Copy Generator and the Trickster Generator, as shown in Figure~\ref{Adv}.

\textbf{Deviation Copy Generator.}
The Deviation Copy Generator (DevCopy) produces hard positive samples ${\mathcal{X}}_*^+$ that are semantically consistent with the anchor embedding $\mathcal{E}_k^{\prime}$ while being distant in the latent space. This forces the model to capture high-level semantics. The process involves two stages:

 A perturbation $\psi$ is applied to $\mathcal{E}_k^{\prime}$ to minimize the contrastive loss:
   \begin{equation}
   \check{\mathcal{X}} = \mathcal{E}_k^{\prime} - \psi \frac{\boldsymbol{v}}{\|\boldsymbol{v}\|_2}, \quad 
   \boldsymbol{v} = \nabla_{\mathcal{E}_k^{\prime}} \mathcal{L}_{\mathrm{cont}}.
   \end{equation}

To ensure semantic consistency, the KL divergence between the perturbed embedding $\check{\mathcal{X}}$ and the anchor embedding $\mathcal{E}_k^{\prime}$ is minimized:
   \begin{equation}
   \mathcal{L}_{K L} = \sum_{k=1}^K D_{K L}\left(P\left(\mathcal{S}_k \mid \check{\mathcal{X}}\right) \| P\left(\mathcal{S}_k \mid \mathcal{E}_k^{\prime}\right)\right).
   \end{equation}

   The refined positive sample ${\mathcal{X}}_*^+$ is then obtained as:
   \begin{equation}
   {\mathcal{X}}_*^+ = \check{\mathcal{X}} - \psi \frac{\boldsymbol{\omega}}{\|\boldsymbol{\omega}\|_2}, \quad 
   \boldsymbol{\omega} = \nabla_{\check{\mathcal{X}}} \mathcal{L}_{K L}.
   \end{equation}

This two-stage strategy generates ${\mathcal{X}}_*^+$ as a semantically robust and challenging positive sample, enhancing the model’s ability to learn meaningful representations.

\textbf{Trickster Generator.}  
The Trickster Generator creates hard negative samples ${\mathcal{X}}_*^-$ that are structurally similar to the anchor embedding ${\mathcal{E}_k^{\prime}}$ but semantically distinct. This ensures the model can effectively discern fine-grained semantic differences.

To generate ${\mathcal{X}}_*^-$, a small perturbation $\delta$ constrained by a norm limit $\epsilon$ is introduced:
\begin{equation}
\|\delta\|_2 \leq \epsilon.
\end{equation}
The perturbation is optimized to maximize semantic divergence by minimizing the conditional likelihood with respect to the similarity matrix $\mathcal{S}_*$:
\begin{equation}
\mathcal{L}_{\mathrm{neg}} = - \sum_{k=1}^K P(\mathcal{S}_k \mid {\mathcal{X}}_*^-).
\end{equation}

Using a gradient-based approximation, the negative sample ${\mathcal{X}}_*^-$ is calculated as:
\begin{equation}\label{trickster}
{\mathcal{X}}_*^- = {\mathcal{E}_k^{\prime}} - \epsilon \frac{\boldsymbol{g}}{\|\boldsymbol{g}\|_2}, \quad 
\boldsymbol{g} = \nabla_{\mathcal{E}_k^{\prime}} \log p\left(\mathcal{S}_k \mid \mathcal{E}_k\right).
\end{equation}

The generated ${\mathcal{X}}_*^-$ acts as a semantically challenging negative sample, effectively improving the model’s robustness.

\textbf{Integration and Task Relevance.}
To further enhance representation quality, separate linear projections are applied to ${\mathcal{X}}_*^+$ and ${\mathcal{X}}_*^-$ to align them with downstream tasks, inspired by techniques in \cite{TPAMICLSV}. This design is particularly relevant to urban applications such as traffic flow prediction, where robust embeddings can effectively capture complex spatial and temporal dependencies.

\textbf{Algorithm Analysis.}
The pseudo-code for the Deviation Copy Generator and Trickster Generator is presented in Algorithm~\ref{alg:dev} and Algorithm~\ref{alg:tri}, respectively. The time complexity of these algorithms is derived from the core operations within their procedures. For the Deviation Copy Generator, the first stage involves computing the gradient, $\boldsymbol{v} = \nabla_{\mathcal{E}_k^{\prime}} \mathcal{L}_{\mathrm{cont}}$, and performing normalization, both with a complexity of $\mathcal{O}(d)$, where $d$ is the dimensionality of the embedding $\mathcal{E}_k^{\prime}$. The second stage requires computing the KL divergence across $K$ relations, which adds a complexity of $\mathcal{O}(Kd)$, and an additional gradient computation, $\boldsymbol{\omega} = \nabla_{\check{\mathcal{X}}} \mathcal{L}_{K L}$, with a complexity of $\mathcal{O}(d)$. As a result, the total complexity of the Deviation Copy Generator is $\mathcal{O}(Kd)$. 

For the Trickster Generator, the operations involve computing the gradient, $\boldsymbol{g} = \nabla_{\mathcal{E}_k^{\prime}} \log p(\mathcal{S}_k \mid \mathcal{E}_k)$, and performing normalization, both of which require $\mathcal{O}(d)$. Therefore, the overall complexity of the Trickster Generator is $\mathcal{O}(d)$. 

When both generators are applied to $N$ samples, the combined complexity becomes $\mathcal{O}(N \cdot Kd)$ for the Deviation Copy Generator and $\mathcal{O}(N \cdot d)$ for the Trickster Generator. The dominant term in the overall complexity depends on the number of relations $K$, the number of samples $N$, and the embedding dimensionality $d$.

\begin{algorithm}[htbp]
\caption{Deviation Copy Generator (DevCopy)}
\label{alg:dev}
\KwIn{Anchor embedding $\mathcal{E}_k^{\prime}$, contrastive loss $\mathcal{L}_{\mathrm{cont}}$, perturbation step size $\psi$.}
\KwOut{Hard positive sample ${\mathcal{X}}_*^+$.}

\textbf{Stage 1: Initial Perturbation} \\
Compute gradient of contrastive loss:
\[
\boldsymbol{v} = \nabla_{\mathcal{E}_k^{\prime}} \mathcal{L}_{\mathrm{cont}}
\] 

Apply perturbation to the anchor embedding:
\[
\check{\mathcal{X}} = \mathcal{E}_k^{\prime} - \psi \frac{\boldsymbol{v}}{\|\boldsymbol{v}\|_2}
\]

\textbf{Stage 2: Semantic Refinement} \\
Minimize KL divergence between $\check{\mathcal{X}}$ and $\mathcal{E}_k^{\prime}$:
\[
\mathcal{L}_{K L} = \sum_{k=1}^K D_{K L}\left(P\left(\mathcal{S}_k \mid \check{\mathcal{X}}\right) \| P\left(\mathcal{S}_k \mid \mathcal{E}_k^{\prime}\right)\right)
\]

Compute gradient of KL loss:
\[
\boldsymbol{\omega} = \nabla_{\check{\mathcal{X}}} \mathcal{L}_{K L}
\]

Refine the positive sample:
\[
{\mathcal{X}}_*^+ = \check{\mathcal{X}} - \psi \frac{\boldsymbol{\omega}}{\|\boldsymbol{\omega}\|_2}
\]

\Return ${\mathcal{X}}_*^+$
\end{algorithm}

\begin{algorithm}[htbp]
\caption{Trickster Generator}
\label{alg:tri}
\KwIn{Anchor embedding $\mathcal{E}_k^{\prime}$, similarity matrix $\mathcal{S}_*$, perturbation constraint $\epsilon$.}
\KwOut{Hard negative sample ${\mathcal{X}}_*^-$.}

Compute gradient of log-likelihood with respect to anchor embedding:
\[
\boldsymbol{g} = \nabla_{\mathcal{E}_k^{\prime}} \log p\left(\mathcal{S}_k \mid \mathcal{E}_k\right)
\]

Generate hard negative sample:
\[
{\mathcal{X}}_*^- = \mathcal{E}_k^{\prime} - \epsilon \frac{\boldsymbol{g}}{\|\boldsymbol{g}\|_2}
\]

\Return ${\mathcal{X}}_*^-$
\end{algorithm}

\subsubsection{Contrastive Optimization for Robust Augmentation}
To enhance the robustness and reliability of contrastive learning in complex and noisy environments, we employ the InfoNCE Loss \cite{oord2019representation} to simultaneously maximize the similarity between anchor embeddings and their corresponding DevCopy samples while minimizing the similarity with Trickster samples. This dual objective is specifically designed to address the challenges of semantic biases and noisy data in urban profiling tasks. The respective positive and negative contrastive losses are expressed as:
\begin{equation}\label{cont-pos}
\mathcal{L}_{\text {cl}^+}=\sum_{k=1}^K \log \frac{\varphi\left({\mathcal{E}_k^{\prime}}, {\mathcal{X}}_*^+\right)}{\sum_{\boldsymbol{e}_x^{\prime} \in Z^{\prime}} \boldsymbol{\varphi}\left({\mathcal{E}_k^{\prime}}, \boldsymbol{e}_x^{\prime}\right)},
\end{equation}
\begin{equation}\label{cont-neg}
\mathcal{L}_{\text {cl}^-}=\sum_{k=1}^K \log \frac{\varphi\left({\mathcal{E}_k^{\prime}}, {\mathcal{X}}_*\right)}{\sum_{\boldsymbol{e}_x^{\prime} \in Z^{\prime}} \varphi\left({\mathcal{E}_k^{\prime}}, \boldsymbol{e}_x^{\prime}\right)},
\end{equation}
where $\varphi(\cdot)=\exp (\operatorname{cosine}(\cdot) / \tau)$ is a similarity metric scaled by temperature parameter $\tau$, and $Z^{\prime}$ represents the extended set of negative samples including $Z \cup\left\{{\mathcal{X}}_*^-\right\}$. This design encourages the model to focus on high-level semantic features while reducing the influence of spurious correlations or noisy samples. 

To balance the contributions of positive and negative pairs, the total contrastive loss is defined as:
\begin{equation}\label{assl}
\mathcal{L}_{\text{acl}}=\alpha \mathcal{L}_{\text{cl}^-}+(1-\alpha) \mathcal{L}_{\text{cl}^+},
\end{equation}
where $\alpha$ is a tunable hyper-parameter that controls the weighting of negative and positive losses, allowing for dynamic adjustments tailored to specific security-driven applications.

\subsection{Joint Optimization}
To achieve secure and robust learning while maintaining computational efficiency, we propose a joint optimization framework that combines the attentive supervised loss $\mathcal{L}_{\text{asl}}$ with the adversarial contrastive loss $\mathcal{L}_{\text{acl}}$. Additionally, $L_2$ regularization is integrated to constrain the trainable parameters, promoting generalization and stability. The total loss function is formulated as:
\begin{equation}
\label{totalLoss}
\mathcal{L}_{\text {total}}=\beta\mathcal{L}_{\text {asl }}+(1-\beta)\mathcal{L}_{\text {acl }}+\mu\|\Theta\|^2,
\end{equation}
where $\Theta$ denotes the model parameters, $\beta$ controls the balance between supervised and self-supervised components, and $\mu$ regulates the regularization strength.

\section{Experiments} In this section, we present an extensive set of experiments designed to evaluate the effectiveness, security, and reliability of the proposed model.

\subsection{Datasets} The experiments are conducted using several real-world datasets obtained from the NYC Open Data platform \footnote{https://opendata.cityofnewyork.us/}. Specifically, we focus on the borough of Manhattan, New York City, which is commonly used as a benchmark in urban studies. The study area is divided into 180 regions to allow for a detailed spatial analysis. Our datasets include diverse and authentic sources, such as census block shapefiles, taxi trip data, points of interest (POI) data, and check-in data, all of which are sourced from the renowned NYC Open Data platform. These datasets are carefully selected to ensure that our model addresses key challenges related to data security, reliability, and robustness in urban data analysis. A detailed description of each dataset can be found in Table~\ref{table1}.

\begin{table}[!t]
\caption{\rm{The details for the dataset.}}
\centering
\scriptsize
\scalebox{1}{
\begin{tabular}{c|c}
\hline
Dataset                        & Details                               \\ \hline
\multirow{2}{*}{Census blocks} & {\multirow{2}{*}{\begin{tabular}[c]{@{}l@{}}The neighborhoods of Manhattan are divided by road grid \\ into 180 blocks that provide relatively fine-grained areas.\end{tabular}}} \\
                               &  \\ \hline
\multirow{2}{*}{Taxi trips}    & \multirow{2}{*}{\begin{tabular}[c]{@{}c@{}}In the studied areas, about 10 million \\ taxicab trips were recorded in a month.\end{tabular}}                                                          \\
                               &      \\ \hline
\multirow{2}{*}{POI data}      & \multirow{2}{*}{\begin{tabular}[c]{@{}c@{}}Approximately 20,000 PoI locations \\ in 13 categories in the studied region\end{tabular}}                                                               \\
                               &        \\ \hline
\multirow{2}{*}{Check-in data} & \multirow{2}{*}{\begin{tabular}[c]{@{}c@{}}More than 100,000 check-in points with \\ over 200 fine-grained category categories\end{tabular}}                                                        \\
                               &           \\ \hline
\multirow{2}{*}{Crime data}    & \multirow{2}{*}{\begin{tabular}[c]{@{}c@{}}In the regions examined, there are \\ approximately 40,000 criminal records in a year.\end{tabular}}                                                     \\
                               &           \\ \hline
\end{tabular}}
\label{table1}
\end{table}

\subsection{Experimental Settings} EUPAS has been implemented using the PyTorch framework. All experiments are conducted on an Intel(R) Xeon(R) CPU E5-2680 v4 hardware platform equipped with an NVIDIA GeForce RTX 3090 Ti-24G GPU. The model is optimized using the Adam optimizer with a learning rate of 0.001.
The number of layers for the heterogeneous GCN is set to 3. In the multi-head self-attention mechanism, we set the number of heads to 4. Key hyperparameters include a standard deviation ${\sigma} = 0.01$, a coefficient $\mathbf{c} = 0.80$, scale factor ${\sigma} = 1$, $\epsilon = 1$, $\alpha = 0.50$, $\beta = 0.15$, and $\tau = 4$.

For the regional clustering task, we use K-means to group the regional embeddings, incorporating land use information from multiple graphs. Figure~\ref{landuse:1} shows the partition of the Manhattan administrative district into 12 regions based on land use, as defined by the community boards \cite{Berg__New}. Correspondingly, we split the study area into 12 clusters. The clustering outcomes are expected to group regions with similar land use types.

\begin{table*}[!t]
\caption{Performance comparison of different approaches for check-in prediction, land usage classification, and crime prediction tasks.
}
\centering
\scalebox{1.1}{
\begin{tabular}{c|ccc|cc|ccc}
\hline
\multirow{2}{*}{Models} & \multicolumn{3}{c|}{Check-in Prediction} & \multicolumn{2}{c|}{Land Usage Classification} & \multicolumn{3}{c}{Crime Prediction} \\ \cline{2-9} 
            & MAE    & RMSE   & $R^2$ & NMI  & ARI  & MAE    & RMSE   & $R^2$ \\ \hline
LINE        & 564.59 & 853.82 & 0.08  & 0.17 & 0.01 & 117.53 & 152.43 & 0.06  \\
GAE         & 498.23 & 803.34 & 0.09  & 0.47 & 0.23 & 96.55  & 133.10 & 0.19  \\
node2vec    & 372.83 & 609.47 & 0.44  & 0.58 & 0.35 & 102.00 & 135.61 & 0.16  \\
GCN         & 489.12 & 765.23 & 0.13  & 0.48 & 0.25 & 96.21  & 134.89 & 0.19  \\
GAT         & 465.21 & 721.86 & 0.42  & 0.35 & 0.28 & 92.67  & 131.23 & 0.22  \\
GraphSage   & 478.00 & 751.22 & 0.23  & 0.29 & 0.19 & 93.58  & 133.11 & 0.21  \\
POI         & 482.12 & 568.21 & 0.39  & 0.31 & 0.16 & 94.71  & 129.01 & 0.24  \\
HDGE        & 399.28 & 536.27 & 0.57  & 0.59 & 0.29 & 72.65  & 96.36  & 0.58  \\
ZE-Mob      & 360.71 & 592.92 & 0.47  & 0.61 & 0.39 & 101.98 & 132.16 & 0.20  \\
MV-PN       & 476.14 & 784.25 & 0.08  & 0.38 & 0.18 & 92.30  & 123.96 & 0.30  \\
CGAL        & 315.58 & 524.98 & 0.59  & 0.69 & 0.45 & 69.59  & 93.49  & 0.60  \\
MVGRE       & 297.72 & 495.27 & 0.63  & 0.78 & 0.59 & 65.16  & 88.19  & 0.64  \\
MGFN        & 280.91 & 436.58 & 0.72  & 0.76 & 0.58 & 59.45  & 77.60  & 0.72  \\
ROMER       & 252.14 & 413.96 & 0.74  & 0.81 & 0.68 & 64.47  & 85.46  & 0.72  \\
HREP        & 270.28 & 406.53 & 0.75  & 0.80 & 0.65 & 65.66  & 84.59  & 0.68  \\
EUPAS(Ours) & 251.70 & 394.68 & 0.77  & 0.84 & 0.71 & 58.56  & 78.41  & 0.73  \\ \hline
\end{tabular}}
\label{metr}
\end{table*}

\subsection{Baselines} This paper compares the EUPAS model with the following baselines:

\begin{itemize} \item {\bfseries LINE} \cite{Tang_2015_LINE}: Preserves both local and global network structures by optimizing a suitable objective function.

\item {\bfseries node2vec} \cite{Grover_2016_node2vec}: Concatenates the embeddings of each graph to obtain region embeddings.
\item {\bfseries GCN} \cite{Kipf_2017_SemiSupervised}: A widely-used graph neural network architecture that leverages convolution-based message passing to aggregate information from neighboring nodes.
\item {\bfseries GraphSage} \cite{GraphSAGE}: Aggregates information from sampled subgraphs to optimize the computational and memory efficiency of graph neural networks.
\item {\bfseries GAE} \cite{Kipf_2016_Variational}: A graph autoencoder that learns node embeddings by optimizing for the reconstruction of input graph data in a latent space.
\item {\bfseries GAT} \cite{Velickovic_2018_GAT}: Introduces attention mechanisms to graph neural networks, allowing for differentiated attention to neighboring nodes based on their relevance during message propagation.

\item {\bfseries POI} \cite{Zhang_2020_MVGRE}: A baseline approach that employs the attributes of POI to represent spatial regions. The TF-IDF algorithm is applied to assess the relevance between different regions based on their POI vector representations.

\item {\bfseries HDGE} \cite{Wang_2017_HDGE}: Uses path sampling on traffic flow and spatial graphs to jointly learn region representations.

\item {\bfseries ZE-Mob} \cite{Yao_2018_ZeMob}: Learns region embeddings by considering the co-currency relation of regions in human mobility trips.

\item {\bfseries MV-PN} \cite{Fu_2019_MVPN}: Learns region embeddings using a region-wise multi-view POI network.

\item {\bfseries CGAL} \cite{CGAL}: Encodes region embeddings using unsupervised methods applied to POI and mobility graphs. Adversarial learning is used to integrate intra-region structures and inter-region dependencies.

\item {\bfseries MVGRE} \cite{Zhang_2020_MVGRE}: Implements cross-view information sharing and weighted multi-view fusion to extract region embeddings based on mobility patterns and inherent region properties (e.g., POI, check-in).

\item {\bfseries MGFN} \cite{Wu_2022_MGFN}: Uses multi-graph fusion networks with a multi-level cross-attention mechanism to learn region embeddings from multiple mobility patterns, integrated through a mobility graph fusion module.

\item {\bfseries ROMER} \cite{ROMER}: Excels in urban region embedding by capturing multi-view dependencies from diverse data sources using global graph attention networks and a two-stage fusion module.

\item {\bfseries HREP} \cite{HREP}: Applies the continuous prompt method prefix-tuning, replacing direct region embedding usage in downstream tasks. This approach provides task-specific guidance, improving overall performance. \end{itemize}

\subsection{Evaluation Metrics}
To evaluate the performance of our model, we use different metrics for various tasks.
For the check-in and crime prediction tasks, we use the following metrics:

\begin{itemize}
    \item \textbf{Mean Absolute Error (MAE)}:
    \begin{equation}
    \text{MAE} = \frac{1}{n} \sum_{i=1}^{n} | y_i - \hat{y}_i |
    \end{equation}
    
    \item \textbf{Root Mean Squared Error (RMSE)}:
    \begin{equation}
    \text{RMSE} = \sqrt{\frac{1}{n} \sum_{i=1}^{n} (y_i - \hat{y}_i)^2}
    \end{equation}

    \item \textbf{Coefficient of Determination ($R^2$)}:
    \begin{equation}
    R^2 = 1 - \frac{\sum_{i=1}^{n} (y_i - \hat{y}_i)^2}{\sum_{i=1}^{n} (y_i - \bar{y})^2}
    \end{equation}
\end{itemize}
where \( y_i \) is the true value, \( \hat{y}_i \) is the predicted value, \( \bar{y} \) is the mean of the true values, and \( n \) is the number of observations.

For the land usage classification task, we employ the following metrics:

\begin{itemize}
    \item \textbf{Normalized Mutual Information (NMI)}:
    \begin{equation}
    \text{NMI}(U,V) = \frac{2I(U;V)}{H(U) + H(V)}
    \end{equation}

    \item \textbf{Adjusted Rand Index (ARI)}:
    \begin{equation}
    \text{ARI} = \frac{\text{RI} - \mathbb{E}[\text{RI}]}{\max(\text{RI}) - \mathbb{E}[\text{RI}]}
    \end{equation}
\end{itemize}
where \( I(U;V) \) is the mutual information between two labelings \( U \) and \( V \), and \( H(U) \) and \( H(V) \) are the entropies of \( U \) and \( V \), respectively. \( \text{RI} \) is the Rand Index, and \( \mathbb{E}[\text{RI}] \) is its expected value.

\begin{figure*}[!t]
\centering
\subfloat[Districts]{
\includegraphics[width=1.05in]{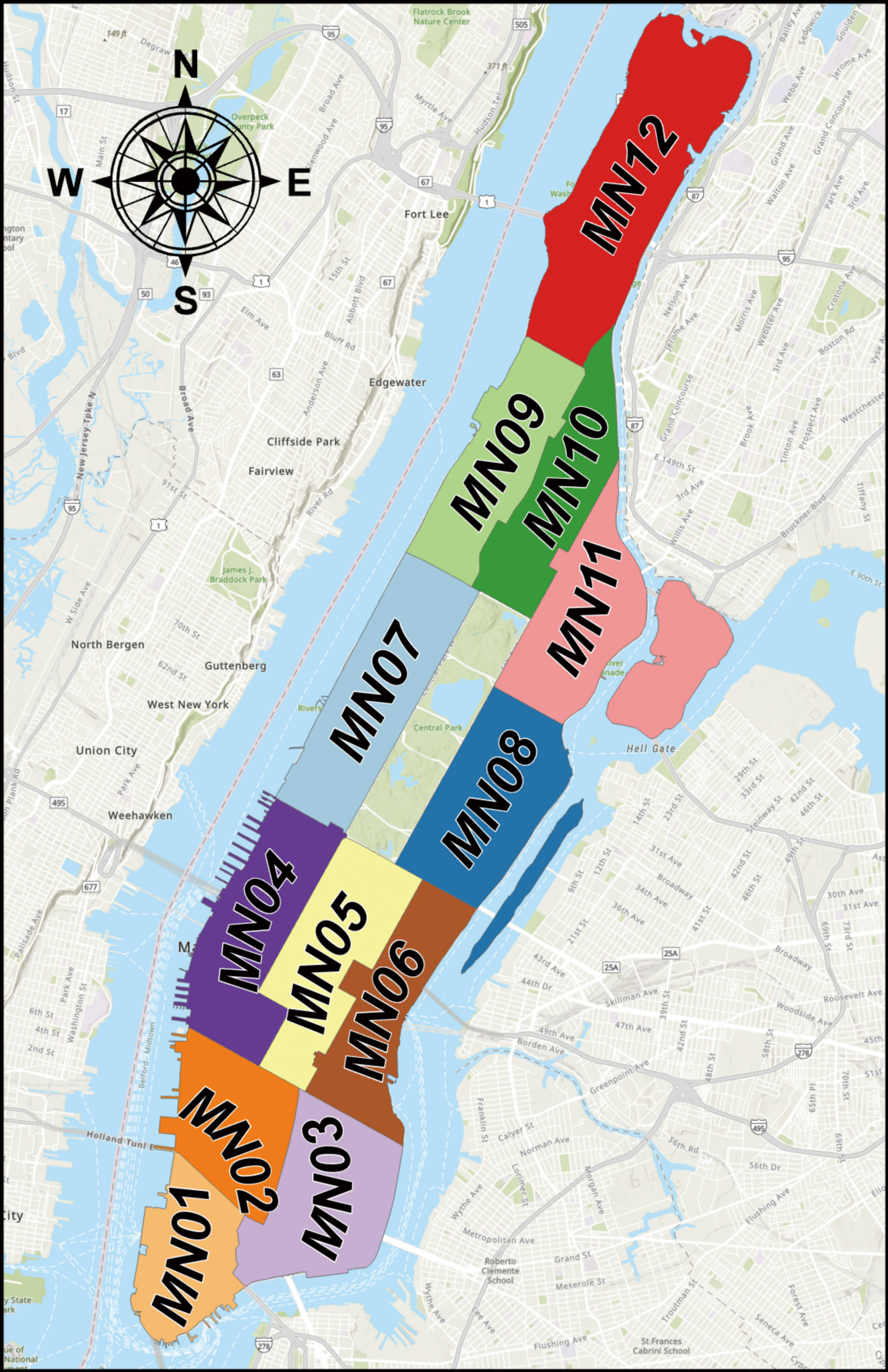} \label{landuse:1}
}
\subfloat[HDGE]{
\includegraphics[width=1.05in]{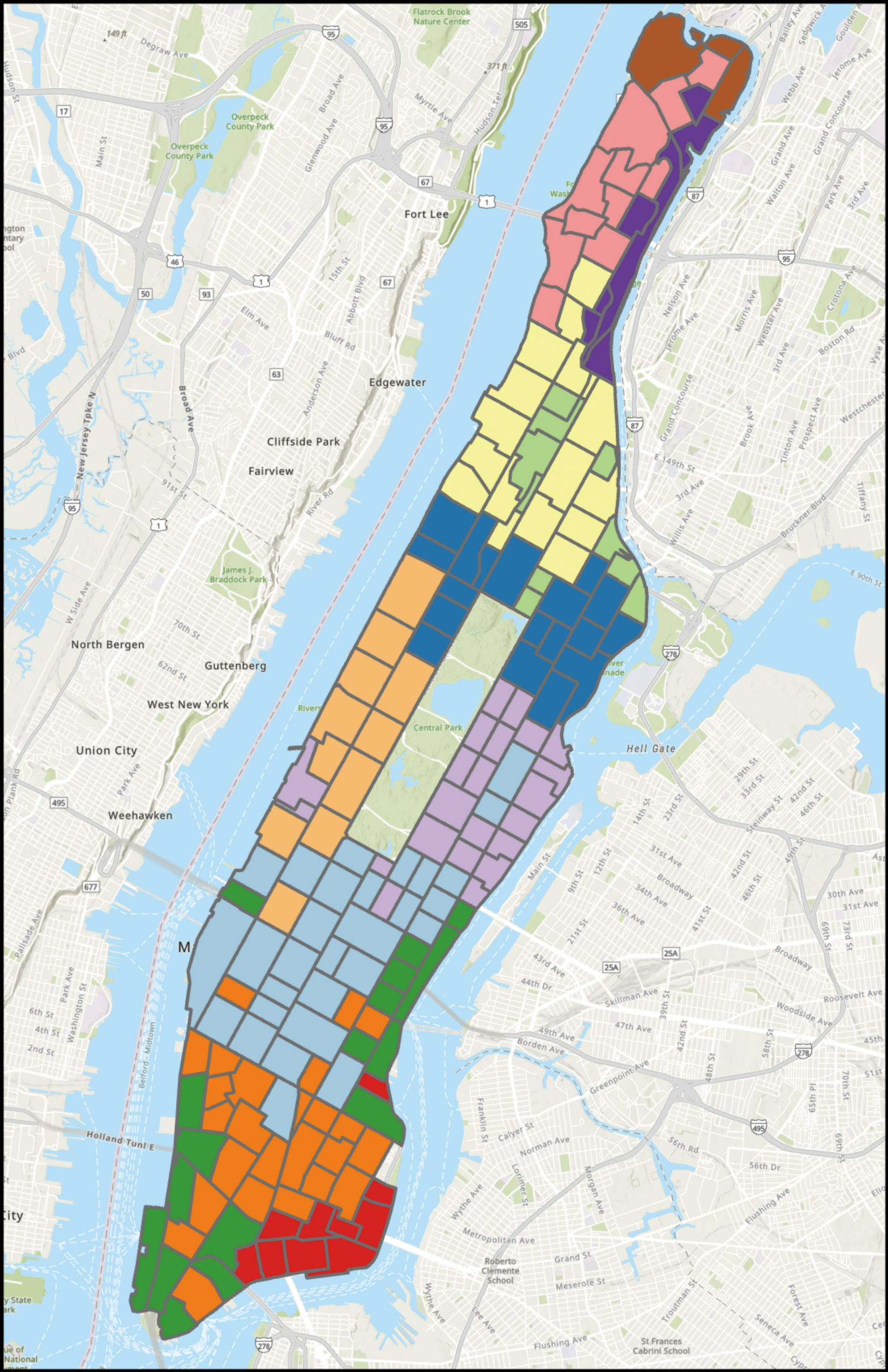}
}
\subfloat[ZE-Mob]{
\includegraphics[width=1.05in]{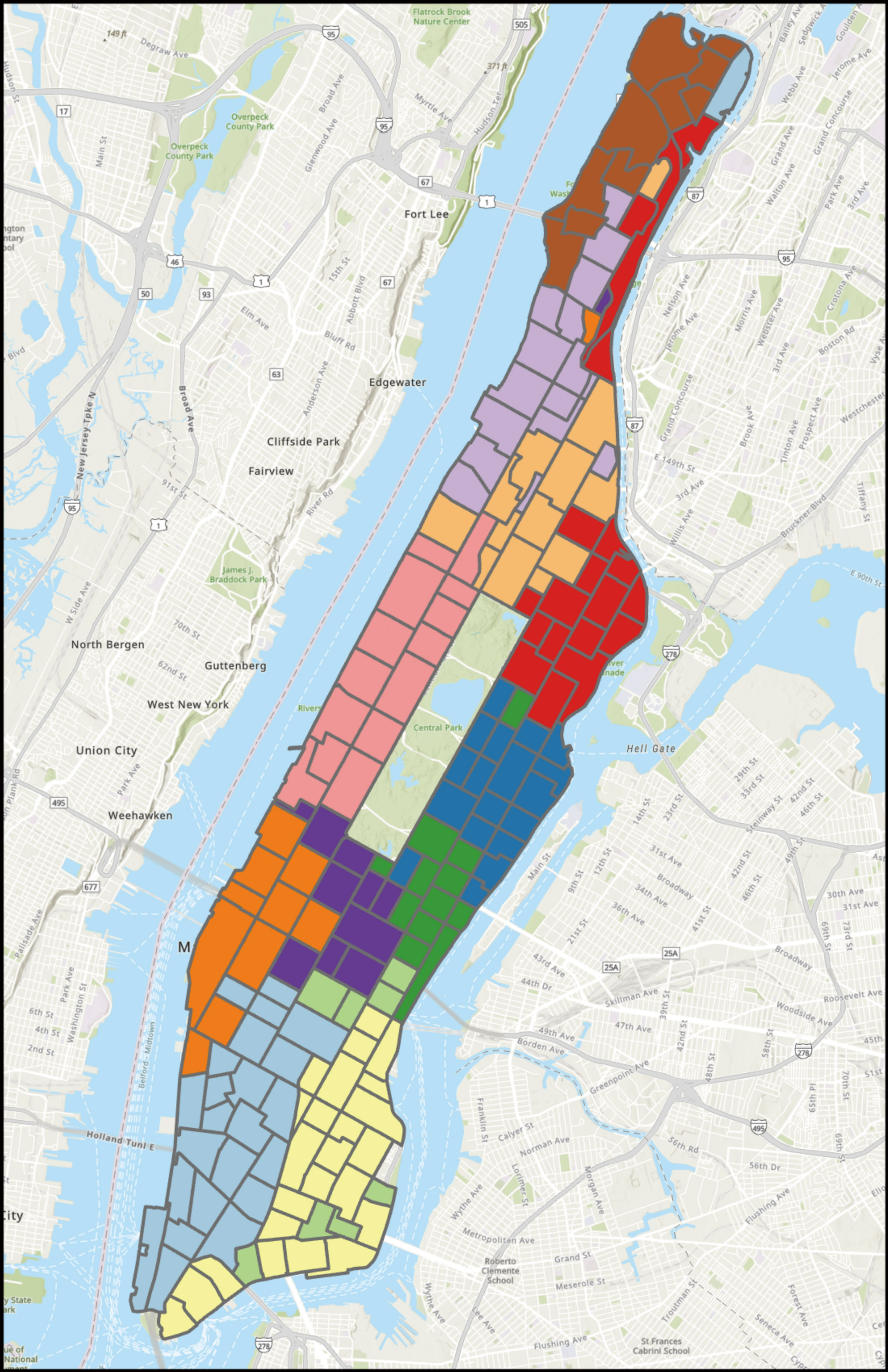}
}
\subfloat[MVGRE]{
\includegraphics[width=1.05in]{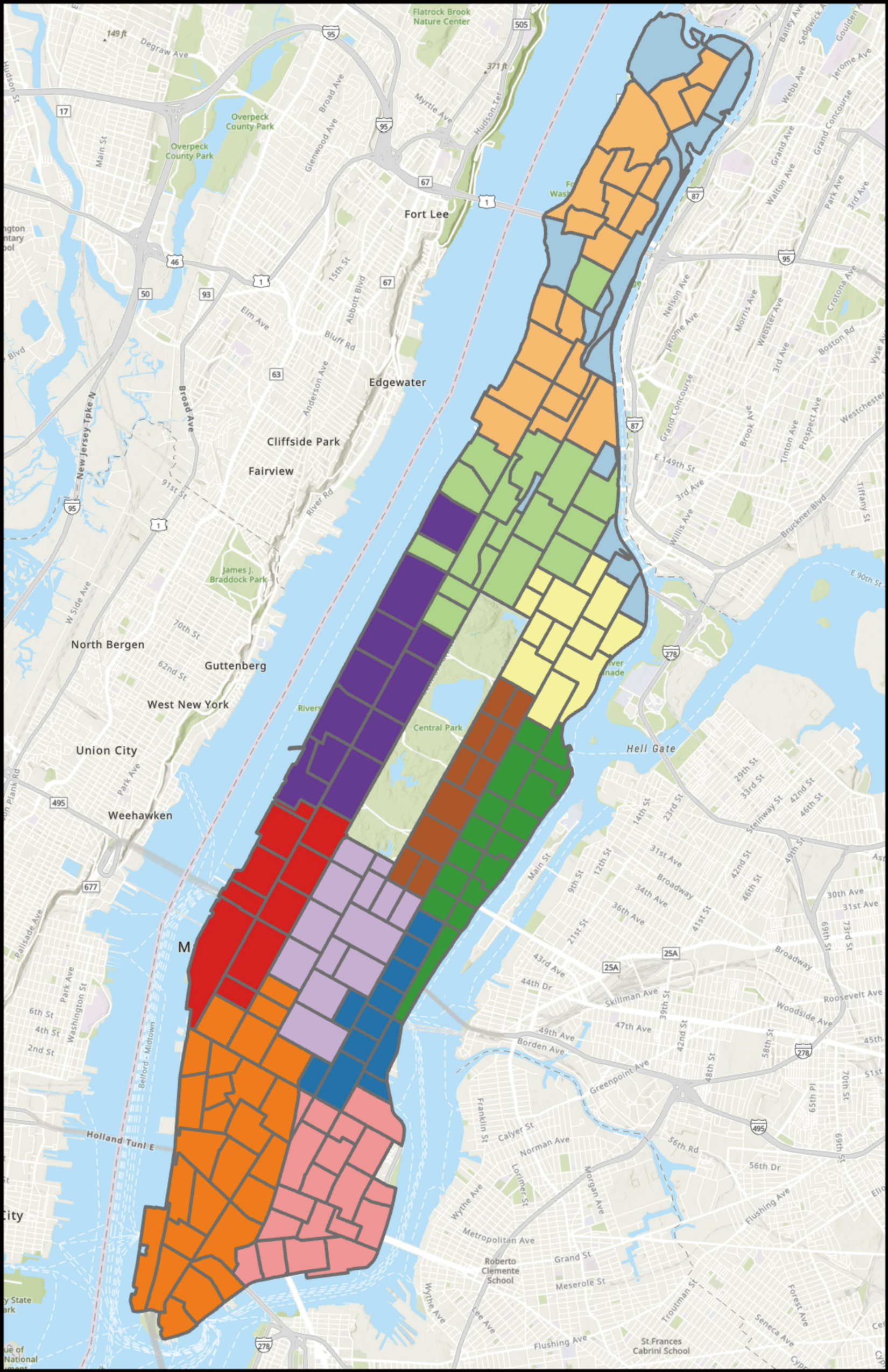}
}
\subfloat[ROMER]{
\includegraphics[width=1.05in]{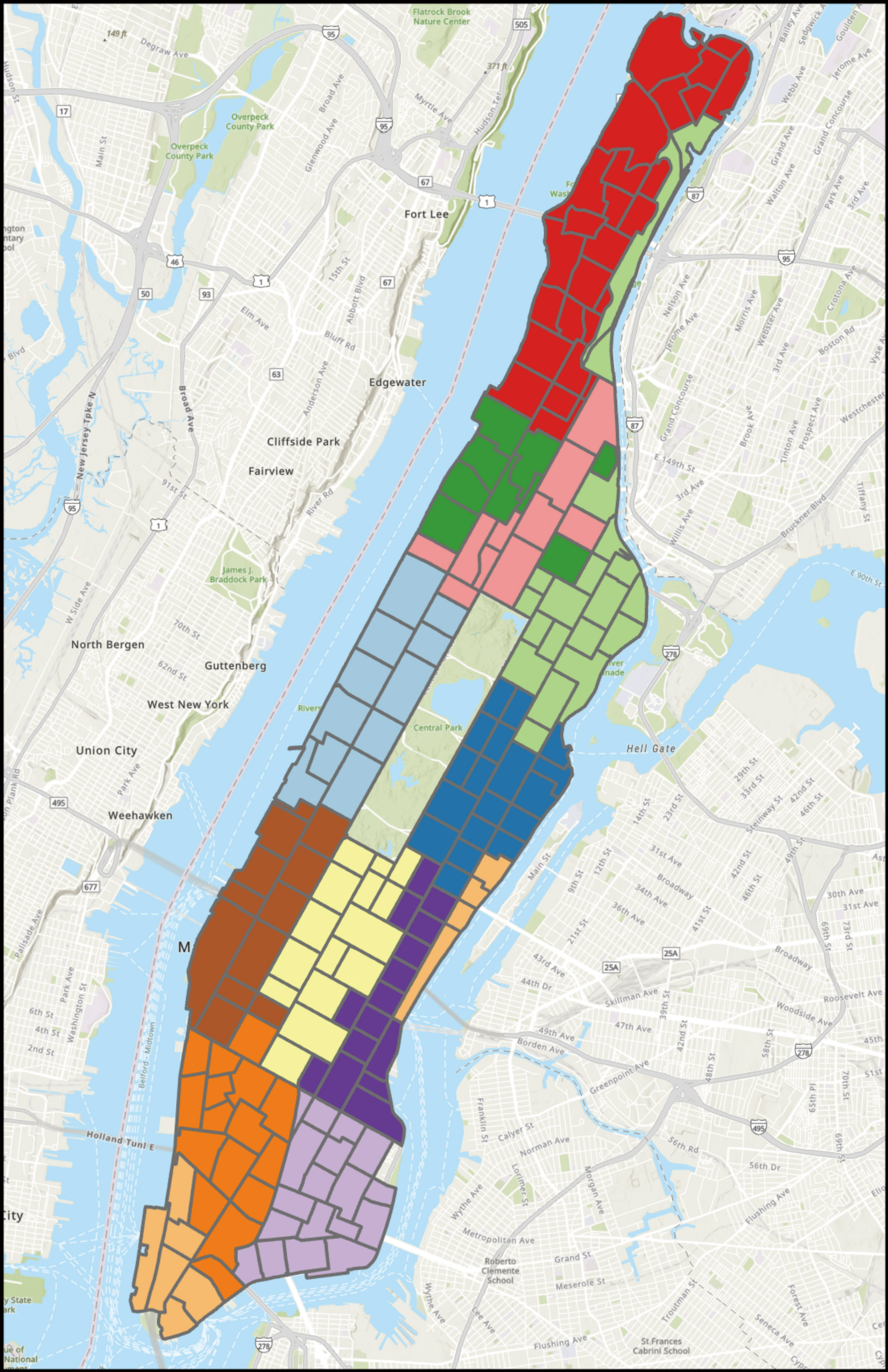}
}
\subfloat[Ours]{
\includegraphics[width=1.05in]{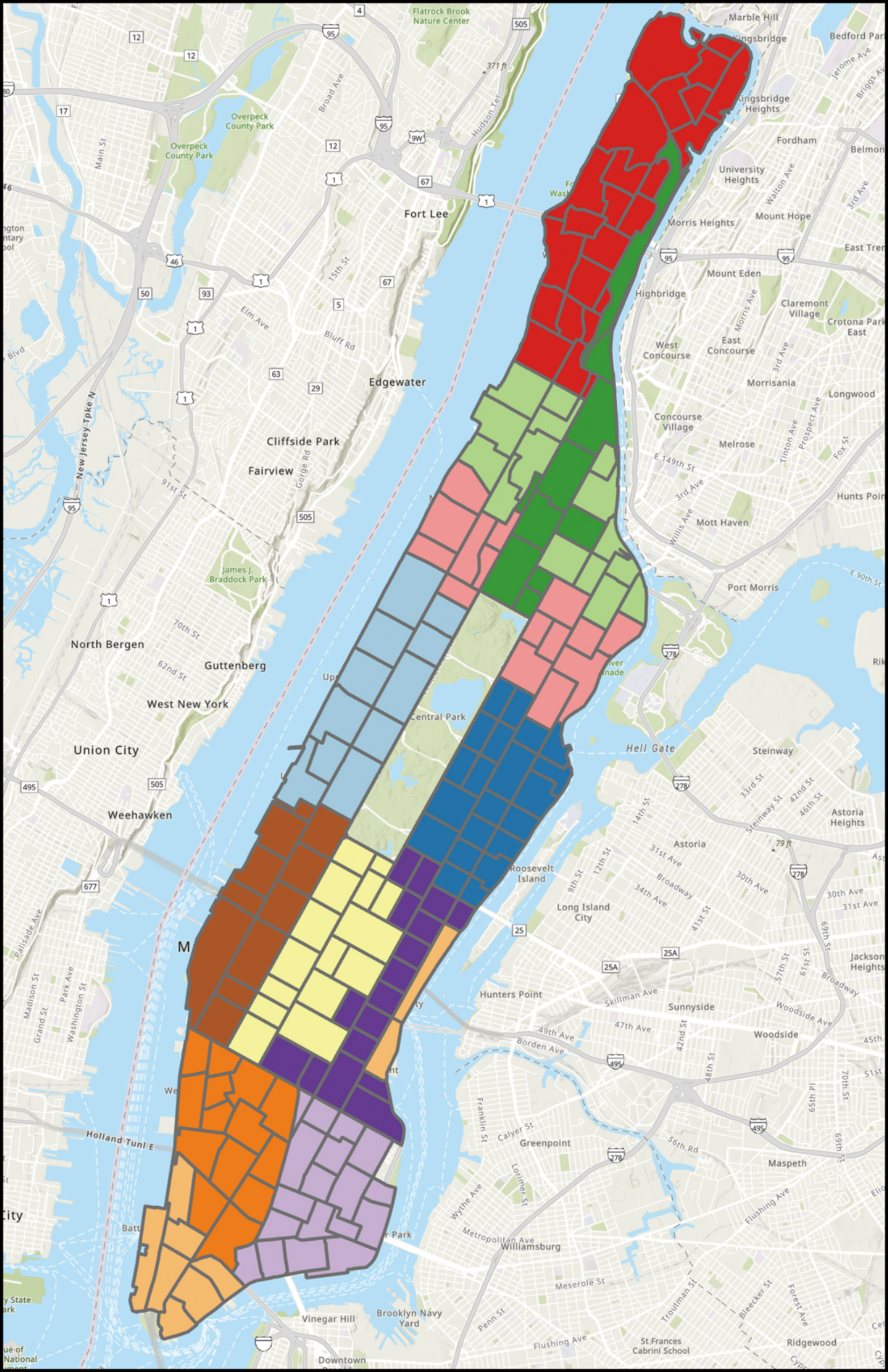}
}
\caption{Comparison of Manhattan districts and region clusters obtained under different baseline methods.}
\label{landuse}
\end{figure*}

\subsection{Main Results}
In this section, we report and analyze the comparison results of the check-in prediction task, the land use classification task, and the crime prediction task. 

\subsubsection{Check-in and Crime Prediction}
In the check-in and crime prediction tasks, we predict the number of crimes and check-in events in each region over the course of one year using the learned region embeddings. We evaluate model performance using metrics such as MAE, RMSE, and $R^2$. The results, shown in Table~\ref{metr}, demonstrate significant improvements over the best baseline, HREP. Specifically, in the check-in prediction task, we achieve a 6.8\% improvement in MAE, a 2.9\% improvement in RMSE, and a 2.6\% improvement in $R^2$. For the crime prediction task, the improvements are 10.8\%, 8.2\%, and 7.3\%, respectively. These results highlight the effectiveness of our joint attentive supervised and contrastive learning framework, which captures complex, nonlinear dependencies and high-level semantic features, crucial for handling noise-related challenges.

In contrast, traditional graph embedding methods like LINE and node2vec perform poorly due to their reliance on local sampling techniques, which fail to capture the relationships between nodes. These models are based on simpler algorithms that cannot model the high-dimensional interdependencies required for accurate urban data modeling, resulting in inferior performance compared to more advanced approaches. Similarly, GAE suffers from its focus on auto-encoding without leveraging external dependencies, limiting its ability to capture the dynamic and noisy nature of urban data.

Deep learning-based methods such as HDGE, ZE-Mob, and MV-PN generally outperform traditional graph embedding techniques by leveraging multi-scale graph structures and advanced embedding methods to capture intricate urban relationships. However, they still struggle with the variability and complexity of real-world urban data due to their simplistic handling of node relationships and the lack of dynamic context-awareness.

Models like MVGRE, MGFN, ROMER, and HRPE enhance performance by incorporating multi-view fusion and attention mechanisms, allowing them to capture node interdependencies more effectively. Despite these advancements, they fail to fully address the challenges posed by noise and incomplete urban region data. These limitations highlight the need for robust methods, such as adversarial training and contrastive learning. EUPAS, with its adversarial self-supervised learning and contrastive learning modules, effectively mitigates these issues. By generating hard positive-negative pairs, the adversarial contrastive module enhances model robustness, making EUPAS more resilient to adversarial attacks and noise, while learning urban region embeddings that are both discriminative and robust to perturbations.

\begin{figure}[!t]
\centering\includegraphics[width=0.465\textwidth]{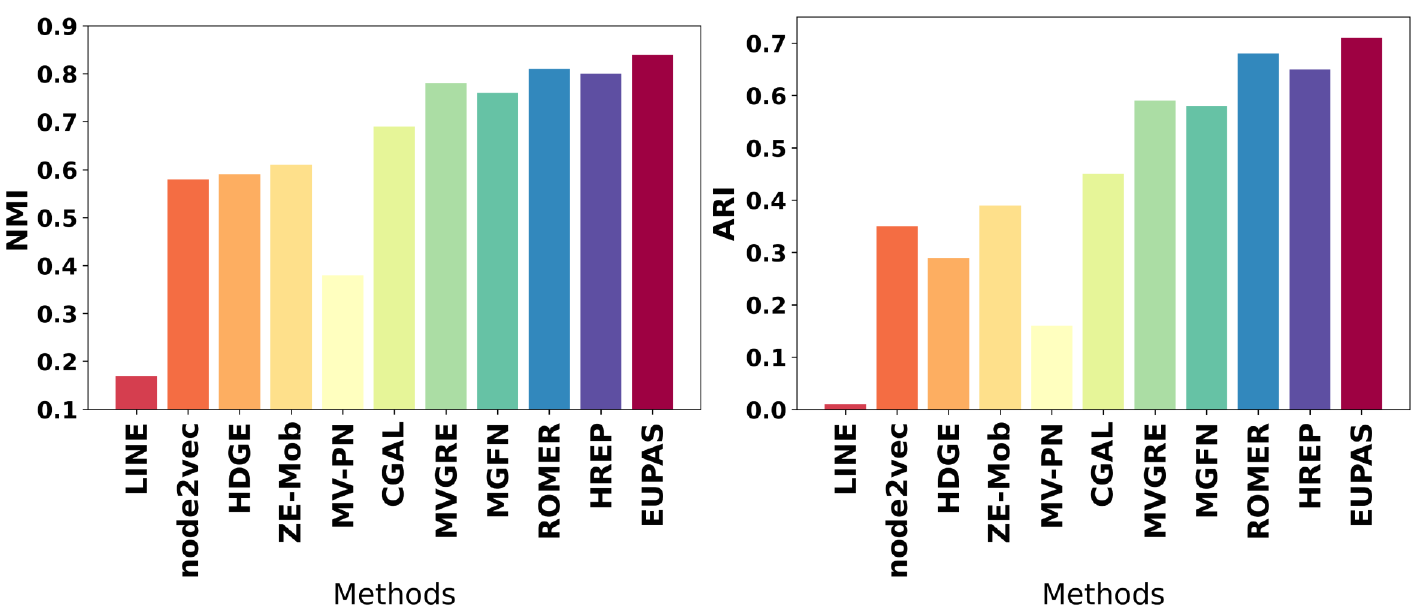}
   \caption{Land usage classification performance.}
   \label{luc}
\end{figure}

\subsubsection{Land Usage Classification}
The results of the land usage classification task are presented in Table~\ref{metr} and Figure \ref{luc} and visualized in Figure~\ref{landuse}. To evaluate the effectiveness of our model, we employ K-means clustering on the region embeddings, and the clustering outcomes are visually represented for intuitive interpretation. As shown in Fig.~\ref{landuse:1}, the community boards \cite{Berg__New} have partitioned the Manhattan administrative district into 12 regions based on land use. Correspondingly, we divide the study area into 12 clusters. The objective of this clustering is to group regions that share similar land use types. 

To quantitatively assess the clustering quality, we adopt Normalized Mutual Information (NMI) and Adjusted Rand Index (ARI), as suggested by \cite{Yao_2018_ZeMob}. Our approach achieves an improvement of more than 3.7\% in NMI and over 4.4\% in ARI compared to the best baseline, ROMER. These results highlight the efficiency and effectiveness of our method in capturing the underlying land use patterns.

For a more intuitive understanding, we provide a visual comparison of the clustering results for five baselines and our model in Fig.~\ref{landuse}. In these visualizations, regions belonging to the same cluster are marked with the same color. Our method's clustering results show a remarkable alignment with the real-world land use boundaries, indicating that the region embeddings learned by our model are more accurate in reflecting regional functions. This is a direct result of our model's ability to effectively exploit high-level semantic features in noisy or incomplete urban data, capture universal human mobility patterns, and generate more precise urban region embeddings.

\begin{figure*}[!t]
\centering
\subfloat[MAE in Check-in Prediction]{
\includegraphics[width=2.2in]{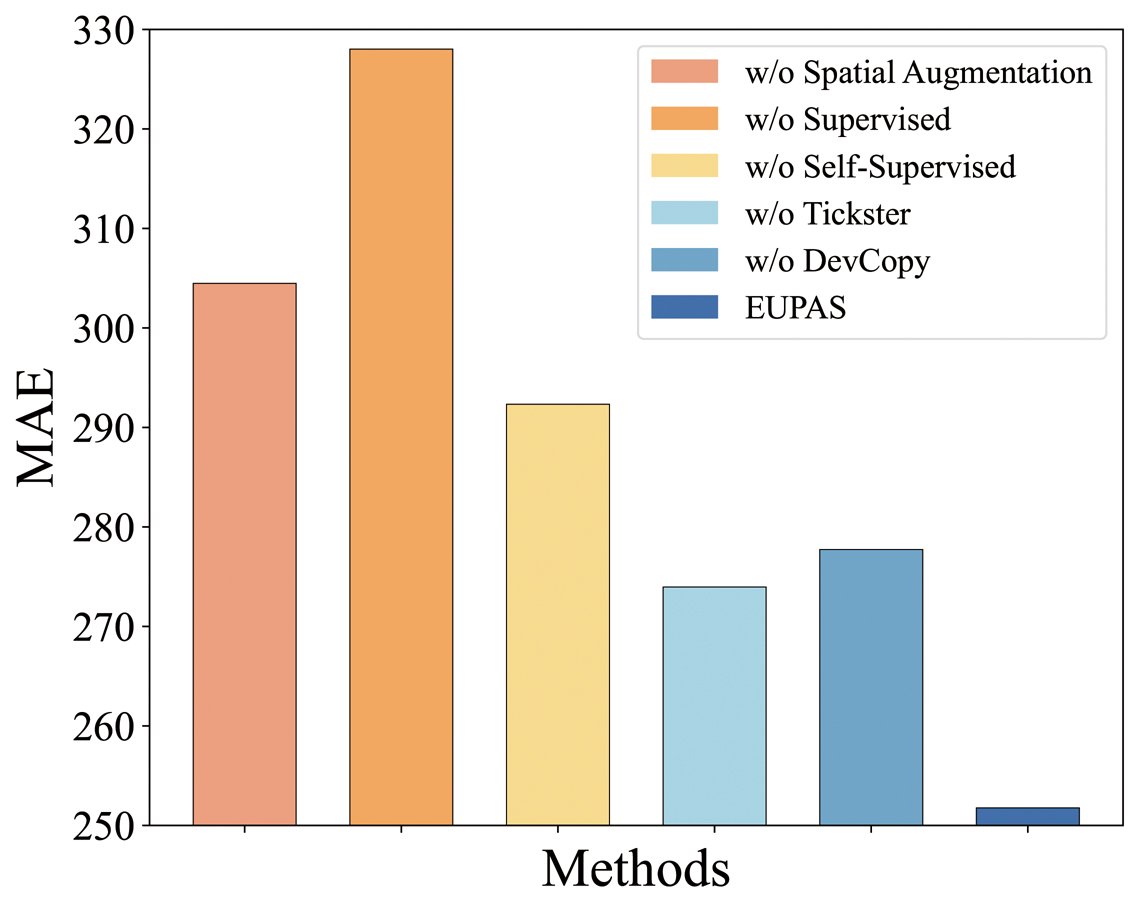}
}
\subfloat[RMSE in Check-in Prediction]{
\includegraphics[width=2.2in]{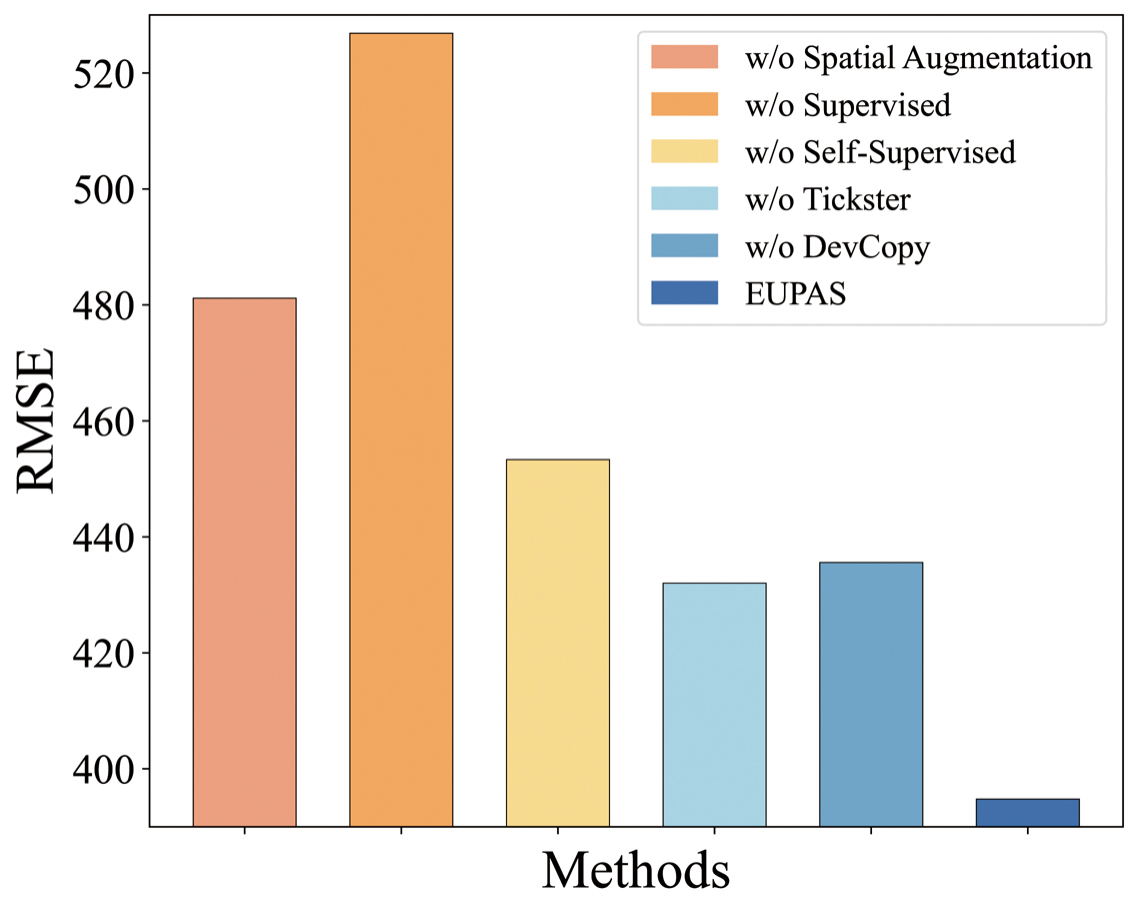}
}
\subfloat[NMI in Land Usage Classification]{
\includegraphics[width=2.2in]{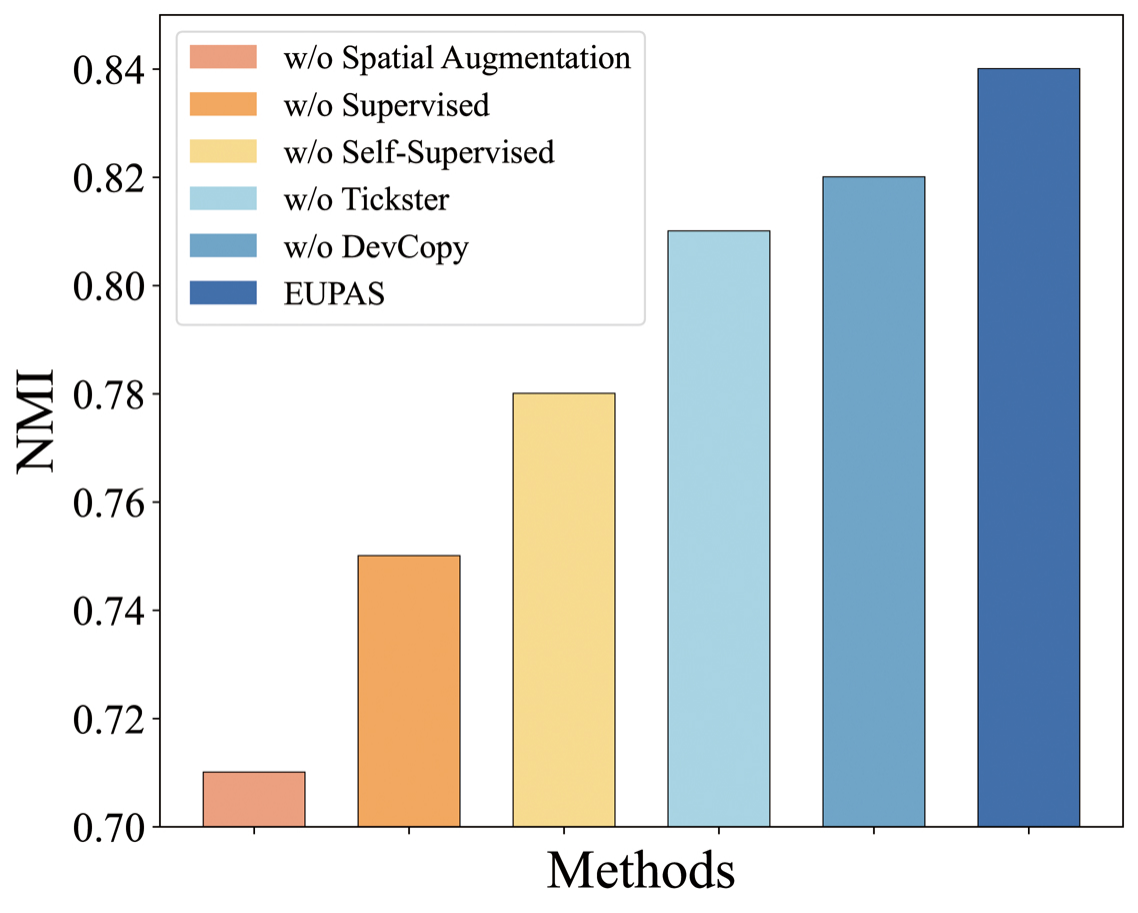}
}
\\
\subfloat[MAE in Crime Prediction]{
\includegraphics[width=2.2in]{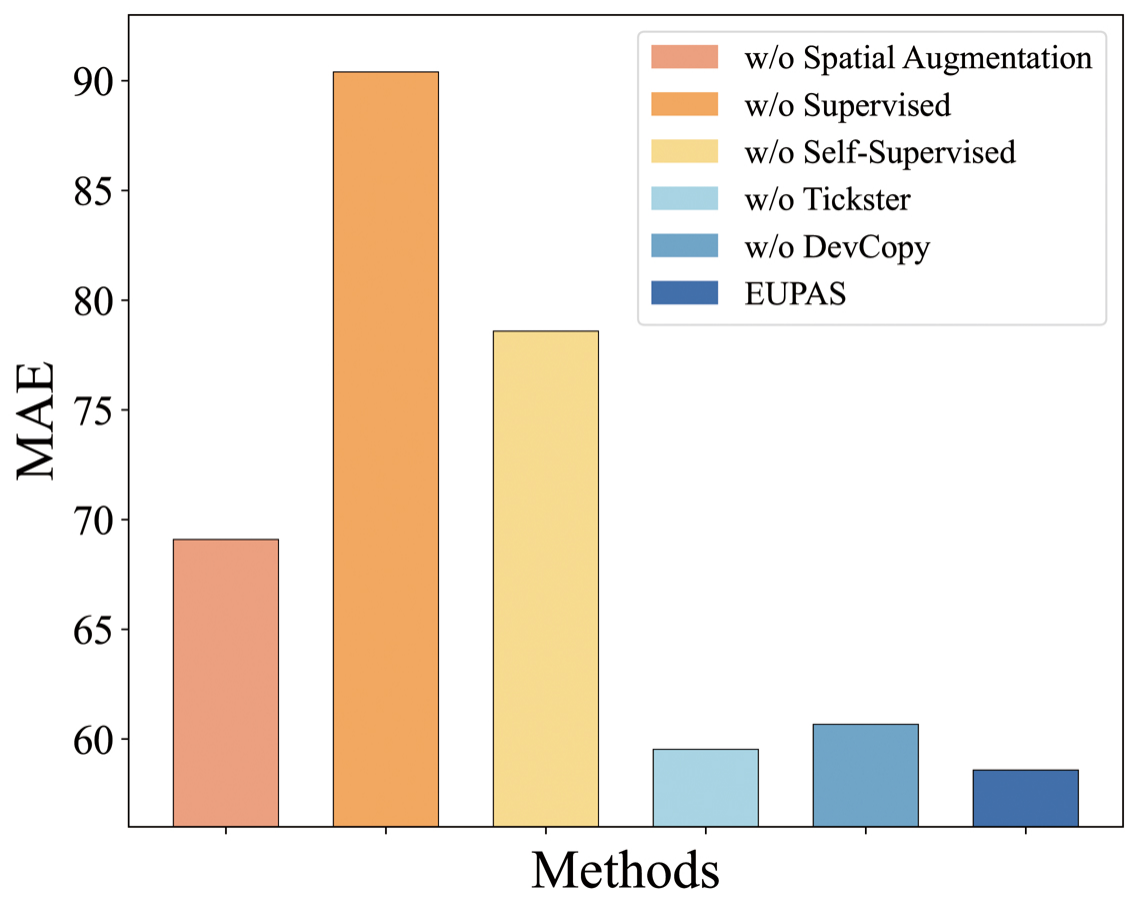}
}
\subfloat[RMSE in Crime Prediction]{
\includegraphics[width=2.2in]{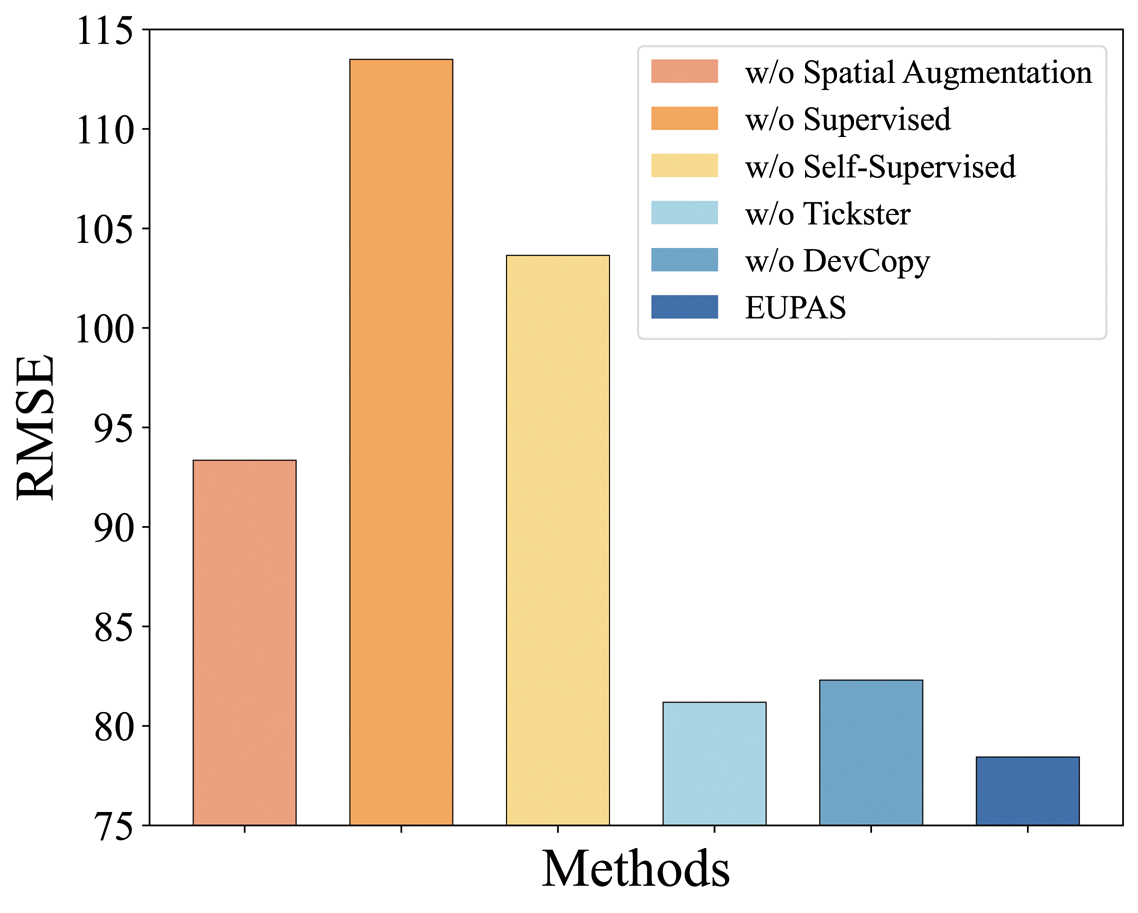}
}
\subfloat[ARI in Land Usage Classification]{
\includegraphics[width=2.2in]{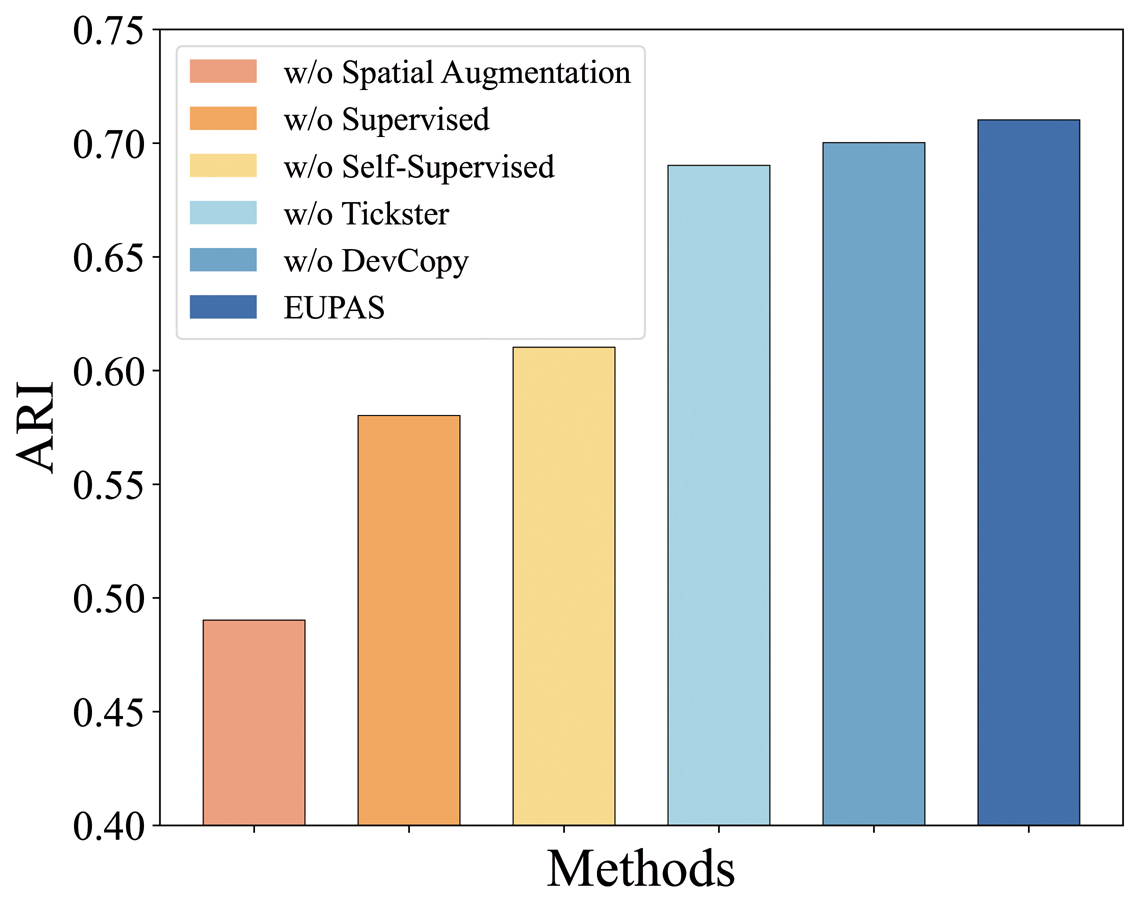}
}
\caption{Ablation studies for three tasks on NYC dataset.}
\label{abla}
\end{figure*}

\subsection{Ablation Study}
To further investigate the contribution of each component, we conduct an ablation study across land usage classification, check-in prediction, and crime prediction tasks. We design the following variants of EUPAS for comparison:

\begin{itemize}
\item \textbf{w/o Spatial Augmentation}: In this variant, we replace the perturbation augmentation layer with a standard data augmentation approach to generate positive samples for region embeddings in the joint learning framework. 
\item \textbf{w/o Supervised}: In this case, we eliminate the attentive supervised module from EUPAS. 
\item \textbf{w/o Self-Supervised}: Here, we remove the adversarial self-supervised module from EUPAS. 
\item \textbf{w/o Tickster}: This variant omits the ${\mathcal{X}}_{Tri}$ component from EUPAS while keeping the other settings identical to the original model. 
\item \textbf{w/o DevCopy}: This variant excludes the ${\mathcal{X}}_{Dev}$ component from EUPAS, with the remaining settings unchanged. 
\end{itemize}

Figure~\ref{abla} presents the ablation results for EUPAS and its variants. From these results, we observe several key findings. First, an effective data augmentation strategy is crucial for graph contrastive learning. Specifically, without our proposed spatial perturbation approach, the performance significantly degrades, highlighting the importance of spatial augmentation for enhancing model robustness. In contrast, removing the adversarial self-supervised module leads to even worse performance, suggesting that adversarial training plays a key role in improving the model's resilience to noisy data.

Moreover, the results for the self-supervised-only variant are notably suboptimal, further validating the necessity of combining supervised and self-supervised learning in our framework. This combination enables the model to learn more effective representations by capturing both high-level semantics and regional dependencies.

Additionally, incorporating the tickster and deviation copy modules further enhances model performance. The dynamic difficulty increase in augmented samples strengthens the model’s discriminative ability over time. Finally, the combination of all these components in our final EUPAS model results in the best overall performance, demonstrating the synergetic effects of each module and its contribution to robust, adaptive, and trustworthy data-driven urban modeling.

\begin{figure}[!t]
\centering
\subfloat[$\tau$]{
\includegraphics[width=3.3in]{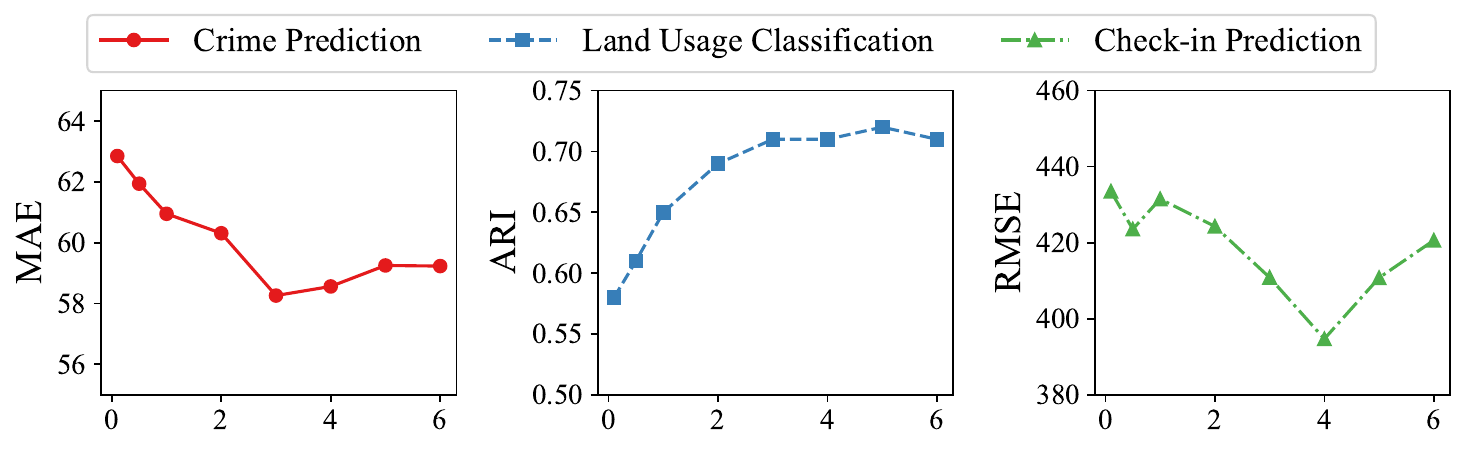} 
\label{tau}
}
\\
\subfloat[$\alpha$]{
\includegraphics[width=3.3in]{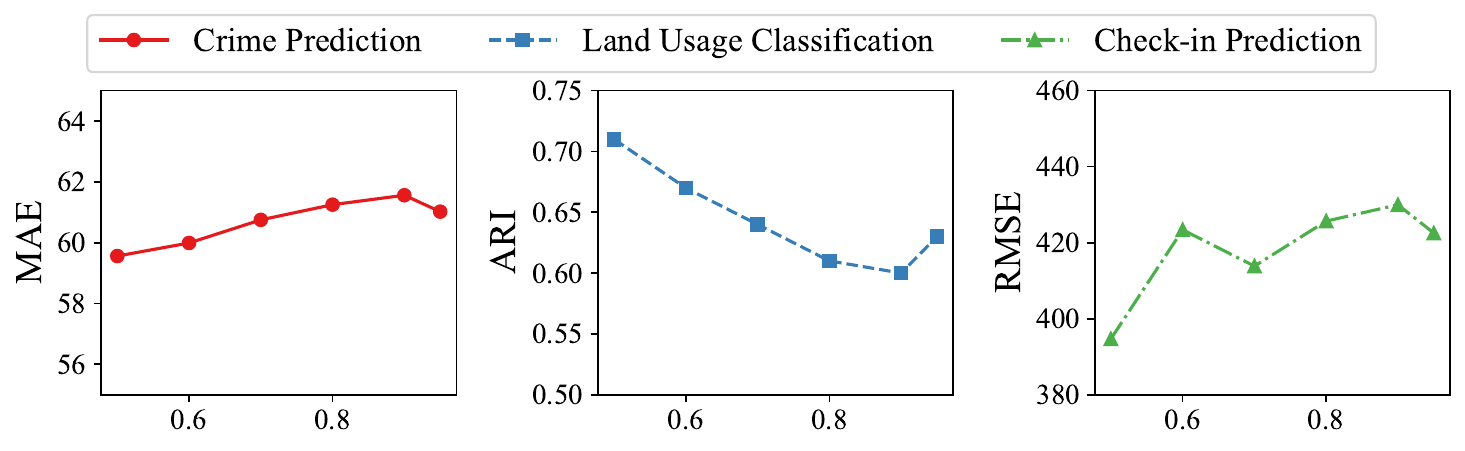}\label{alp}
}
\\
\subfloat[$\beta$]{
\includegraphics[width=3.3in]{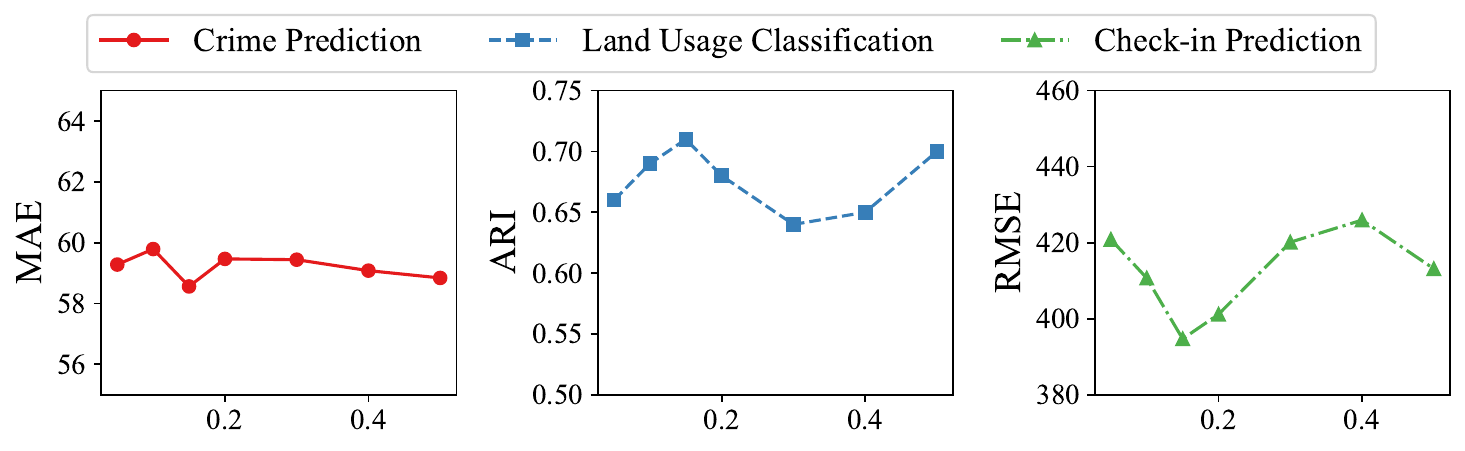}
\label{beta}
}
\caption{Impact of $\tau$ , $\alpha$ and $\beta$ to our model.}
\label{hypter}
\end{figure}

\subsection{Hyperparameter Sensitivity}
In this section, we investigate the impact of different hyperparameter settings on the performance of our EUPAS framework. Specifically, we explore the effects of key hyperparameters, including $\tau$ (used in Equations (\ref{cont-pos}, \ref{cont-neg})), $\alpha$ (Equation (\ref{assl})), and $\beta$ (Equation (\ref{totalLoss})). For each experiment, we systematically vary one hyperparameter while keeping the others fixed at their default values. The experimental results are shown in Figure~\ref{hypter}.

\textbf{The Effect of $\tau$:} The hyperparameter $\tau$ controls the smoothness of embedding similarity in the contrastive loss function. We conduct experiments with values of $\tau$ chosen from the set $\{0.1, 0.5, 1, 2, 3, 4, 5, 6\}$ to assess how it influences performance. As shown in Figure~\ref{tau}, when $\tau \leq 3$, the results are unsatisfactory, likely due to the model overly focusing on challenging negative samples. On the other hand, when $\tau > 4$, the performance gradually deteriorates, as the logit distribution becomes too smooth, causing contrastive learning to treat all negative samples equally and diminishing the model’s ability to discriminate effectively. The optimal value of $\tau$ is found to be 4, striking the right balance between emphasizing challenging negatives and maintaining a smooth similarity distribution for effective contrastive learning.

\textbf{The Effect of $\alpha$:} The hyperparameter $\alpha$ determines the relative weighting of positive and negative sample losses in the adversarial contrastive learning process. We evaluate the performance with $\alpha$ values from the set $\{0.5, 0.6, 0.7, 0.8, 0.9, 0.95\}$. As shown in Figure~\ref{alp}, we observe that $\alpha = 0.5$ produces the best results. This indicates that a balanced weighting of positive and negative samples is essential for effective contrastive learning in our adversarial framework, where $\alpha = 0.5$ ensures that neither positive nor negative samples dominate the learning process.

\textbf{The Effect of $\beta$:} The hyperparameter $\beta$ regulates the relative importance of supervised and self-supervised learning components in the total loss function (Equation~\ref{totalLoss}). We test values of $\beta$ from the set $\{0.05, 0.1, 0.15, 0.2, 0.3, 0.4, 0.5\}$, and the results, shown in Figure~\ref{beta}, suggest that $\beta = 0.15$ is the optimal setting. In our model, supervised learning and self-supervised learning complement each other, with supervised learning having a more significant impact due to its ability to effectively capture high-level region semantics from noisy and incomplete urban data. Setting $\beta = 0.15$ strikes an optimal balance between these components, leading to the best performance across downstream tasks.

\begin{figure}[!t]
\centering\includegraphics[width=0.48\textwidth]{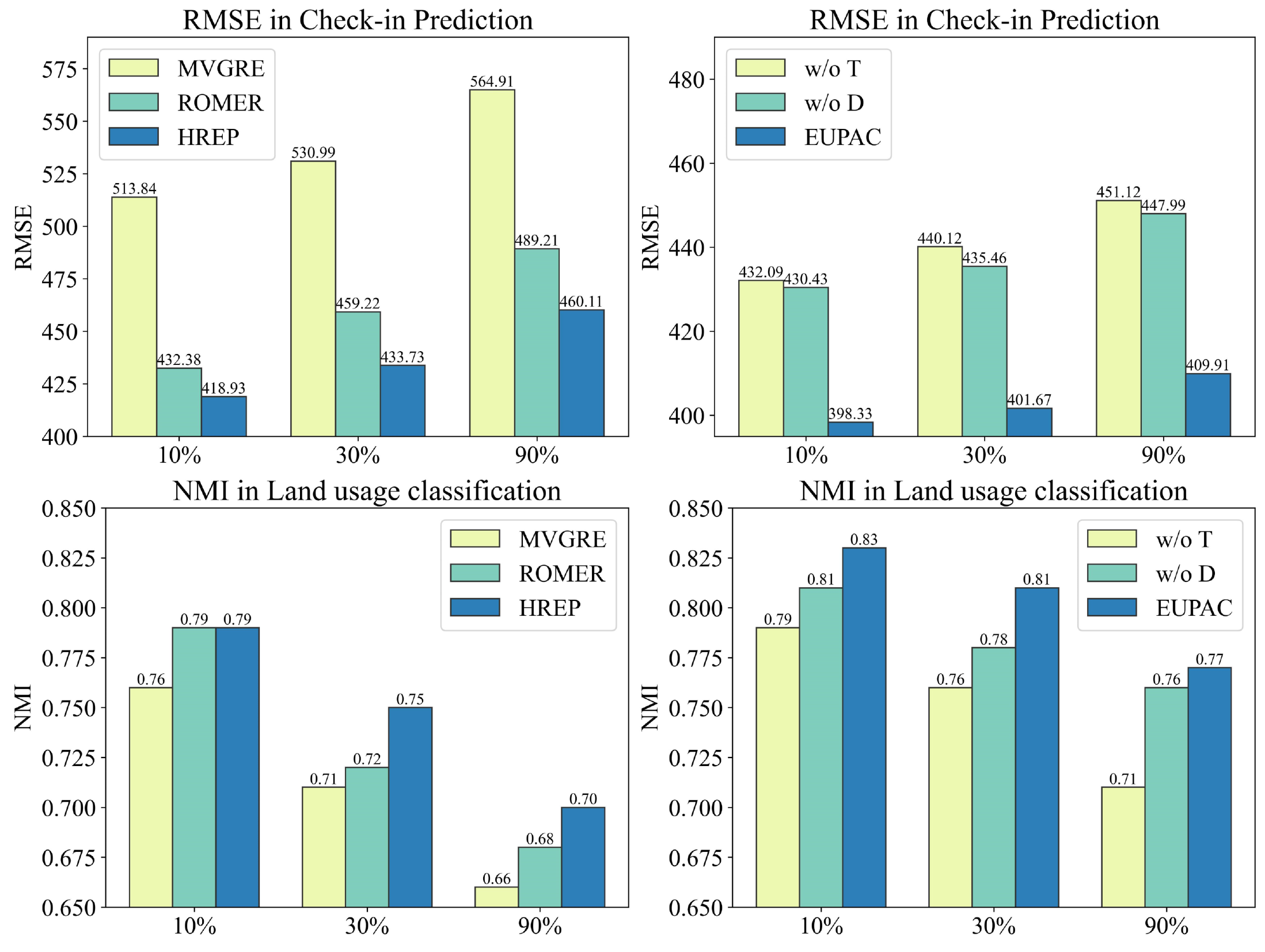}
   \caption{Performance changes as the training data qualities decrease.}
   \label{noise}
\end{figure}

\subsection{Adversarial Robustness Testing}
To assess the robustness of the EUPAS model, we conducted experiments to evaluate its performance under noise and adversarial attacks using two widely used strategies: the Fast Gradient Sign Method (FGSM) \cite{goodfellow2015explaining} and Projected Gradient Descent (PGD) \cite{madry2018towards}.

\subsubsection{Impact of Noise}
We introduced Poisson noise \cite{pasionnoise} to simulate the effect of noise in urban data due to its ability to model count-based or discrete event data, which is common in urban applications (e.g., traffic counts, people movement, or sensor data).  In the experiments, noise sampled from a Poisson distribution (with a noise level of 1) was added to 10\%, 30\%, and 90\% of the input data. The results, shown in Fig.~\ref{noise}, indicate that even in environments with significant noise (e.g., 30\% or more), EUPAS outperforms other methods, maintaining robust performance across downstream tasks. This highlights the strength of the model in handling noisy data. However, it is essential to note that, in real-world applications, we typically expect the noise levels to be lower. Additionally, the ablation study reveals that the Trickster Generator in the adversarial contrastive module shows greater resistance to noise compared to the Deviation Copy Generator.

\subsubsection{Adversarial Attacks}
FGSM \cite{goodfellow2015explaining} generates adversarial examples by perturbing the input data in the direction of the gradient of the loss function with respect to the input. The perturbation $\delta$ is computed as:

\begin{equation}
\delta = \epsilon \cdot \text{sign}(\nabla_x J(\theta, x, y)),
\end{equation}
where $J(\theta, x, y)$ is the loss function, $\theta$ are the model parameters, $x$ is the input, and $y$ is the true label. The parameter $\epsilon$ controls the perturbation magnitude. We applied FGSM to each task and evaluated performance using MAE, RMSE, and $R^2$.

PGD \cite{madry2018towards} is a more iterative and powerful method for generating adversarial examples. It applies FGSM-like perturbations in multiple iterations, with a projection step to ensure the perturbation remains within a specified $\ell_{\infty}$ norm ball around the input. The process is formulated as:

\begin{equation}
x^{t+1} = \text{Proj}_{\mathcal{B}(x, \epsilon)}\left(x^t + \alpha \cdot \text{sign}(\nabla_x J(\theta, x^t, y))\right)
\end{equation}
where $\alpha$ is the step size and $\text{Proj}_{\mathcal{B}(x, \epsilon)}$ projects onto the $\ell_{\infty}$ ball. We iterated for $T$ steps and evaluated performance degradation for each task.

Table~\ref{robustness} presents the performance of EUPAS under FGSM and PGD attacks across three tasks: check-in prediction, land usage classification, and crime prediction. The results indicate a more significant performance drop under PGD due to its iterative nature. In comparison, we also include results from HERP, which lacks a noise resistance mechanism and does not incorporate adversarial defense strategies. As seen in Table~\ref{robustness}, HERP exhibits a significantly larger performance degradation under both FGSM and PGD attacks compared to EUPAS. The lack of noise resistance and adversarial defense leaves HERP vulnerable to perturbations, highlighting the critical need for an integrated approach like ours, which combines adversarial contrastive learning with data augmentation to enhance robustness.

The adversarial contrastive module effectively addresses key challenges in secure and trustworthy data-driven modeling, such as semantic biases, data sparsity, and noise interference. By generating semantically diverse and challenging positive and negative samples, the module enhances the robustness and generalization of learned representations. Furthermore, integrating both supervised and self-supervised objectives within a unified optimization framework ensures the module’s adaptability, making it suitable for security-sensitive applications with varying operational conditions.

The results indicate that while EUPAS performs well in the original setting, adversarial attacks significantly degrade its performance. The PGD attack, being more powerful due to its iterative nature, leads to larger performance drops compared to the FGSM attack. This highlights the vulnerability of the current model to adversarial perturbations, suggesting the need for further research in robustifying the model against such attacks.

\begin{table}[!t]
\centering
\scriptsize
\caption{Adversarial robustness evaluation: Performance comparison of the original and adversarially attacked EUPAS model under FGSM and PGD attacks.}
\scalebox{0.92}{\begin{tabular}{c|c|c|c|c|c}
\hline
Model                  & Task                                                                                  & Metric & Original & FGSM Attack & PGD Attack \\ \hline
\multirow{8}{*}{EUPAS} & \multirow{3}{*}{Check-in Prediction}                                                  & MAE    & 251.70   & 320.58      & 400.92     \\
                       &                                                                                       & RMSE   & 394.68   & 480.74      & 574.12     \\
                       &                                                                                       & $R^2$  & 0.77     & 0.64        & 0.52       \\ \cline{2-6} 
                       & \multirow{2}{*}{\begin{tabular}[c]{@{}c@{}}Land Usage \\ Classification\end{tabular}} & NMI    & 0.84     & 0.79        & 0.73       \\
                       &                                                                                       & ARI    & 0.71     & 0.65        & 0.58       \\ \cline{2-6} 
                       & \multirow{3}{*}{Crime Prediction}                                                     & MAE    & 58.56    & 72.35       & 92.80      \\
                       &                                                                                       & RMSE   & 78.41    & 95.02       & 118.61     \\
                       &                                                                                       & $R^2$  & 0.73     & 0.65        & 0.54       \\ \hline
\multirow{8}{*}{HERP}  & \multirow{3}{*}{Check-in Prediction}                                                  & MAE    & 270.28   & 364.88      & 464.88     \\
                       &                                                                                       & RMSE   & 406.53   & 528.49      & 650.45     \\
                       &                                                                                       & $R^2$  & 0.75     & 0.62        & 0.49       \\ \cline{2-6} 
                       & \multirow{2}{*}{\begin{tabular}[c]{@{}c@{}}Land Usage \\ Classification\end{tabular}} & NMI    & 0.80     & 0.72        & 0.64       \\
                       &                                                                                       & ARI    & 0.65     & 0.59        & 0.52       \\ \cline{2-6} 
                       & \multirow{3}{*}{Crime Prediction}                                                     & MAE    & 65.66    & 85.36       & 105.05     \\
                       &                                                                                       & RMSE   & 84.59    & 109.97      & 135.34     \\
                       &                                                                                       & $R^2$  & 0.68     & 0.58        & 0.48       \\ \hline
\end{tabular}}
\label{robustness}
\end{table}

\begin{table}[!t]
\centering
\caption{Comparison of training time \& inference time.}
\scalebox{0.92}{
\begin{tabular}{@{}cccc@{}}
\toprule
\multirow{2}{*}{Task}                                                                 & \multirow{2}{*}{Model} & \multicolumn{2}{c}{Time (s/100epoch)} \\ \cmidrule(l){3-4} 
 &              & Training & Inference \\ \midrule
\multirow{7}{*}{\begin{tabular}[c]{@{}c@{}}Crime Prediction\end{tabular}} & MVGRE                  & 33.46             & 1.207            \\
 & ROMER        & 31.38                 & 1.232     \\
 & HERP         & 8.143             & 1.987      \\ \cmidrule(l){2-4} 
 & w/o Self-Supervised      & 1.621                 & 1.368     \\
 & w/o Tickster        & 3.703                 & 3.276     \\
 & w/o DevCopy        & 2.713                 & 2.233     \\ \cmidrule(l){2-4} 
 & EUPAS (Ours) &4.024               & 3.812     \\ \bottomrule
\end{tabular}}
\label{time}
\end{table}

\subsection{Computational Efficiency} 
This section compares the training and inference times of recent baselines and EUPAS variants for the crime prediction task. We selected MVGRE, ROMER, and HREP as baseline models because they utilize human mobility data, POI data, and check-in data as inputs, ensuring better control over variables and enabling a more accurate comparison of computational times.

The results are presented in Table~\ref{time}.  Among the baselines, MVGRE and ROMER exhibit the longest training times (33.46s and 31.38s per 100 epochs, respectively), largely due to their simpler yet less efficient structural designs. While HERP achieves a much shorter training time (8.143s), its inference time (1.987s) is notably higher than MVGRE and ROMER, indicating potential inefficiencies in its prediction mechanism.
Among the EUPAS variants, the version without self-supervised learning (w/o Self-Supervised) achieves the fastest training time (1.621s) and a relatively low inference time (1.368s). However, the absence of self-supervised learning sacrifices performance in predictive accuracy and robustness. Similarly, the w/o Tickster variant demonstrates faster training and inference times compared to the full EUPAS model, but its performance is inferior due to the lack of challenging positive-negative pairs generated by the Trickster component. The w/o DevCopy variant strikes a middle ground, with training and inference times of 2.713s and 2.233s, respectively, but also exhibits reduced performance compared to the full model.

The full EUPAS model requires 4.024s for training and 3.812s for inference per 100 epochs. While these times are slightly higher than its variants, the added computational cost is justified by the model's superior performance. EUPAS effectively balances robustness and computational efficiency, making it resilient to adversarial perturbations and noise in urban data.

\section{Conclusion}
In this paper, we propose EUPAS, a robust and efficient framework for urban region representation learning, addressing critical challenges such as noise, data incompleteness, and semantic biases, which are central concerns in the field of secure and trustworthy data-driven modeling. Our framework combines a joint attentive supervised and adversarial contrastive learning approach, which ensures reliable and resilient performance in urban tasks like check-in prediction, crime prediction, and land use classification.
By introducing innovative components such as perturbation augmentation and adversarial contrastive modules, EUPAS effectively mitigates the negative impact of noise and data incompleteness, providing a more robust solution for urban data analysis. 
Our experimental results demonstrate that EUPAS outperforms existing state-of-the-art methods across various metrics, showing superior performance in noisy environments and under adversarial attacks.

Looking forward, we aim to extend EUPAS’s capabilities further by focusing on enhancing its security features. Specifically, we plan to incorporate stronger adversarial defenses to better withstand sophisticated attacks and explore the integration of differential privacy techniques to safeguard sensitive urban data.

\bibliography{EUPAS}

\begin{thebibliography}{10}
\providecommand{\url}[1]{#1}
\csname url@samestyle\endcsname
\providecommand{\newblock}{\relax}
\providecommand{\bibinfo}[2]{#2}
\providecommand{\BIBentrySTDinterwordspacing}{\spaceskip=0pt\relax}
\providecommand{\BIBentryALTinterwordstretchfactor}{4}
\providecommand{\BIBentryALTinterwordspacing}{\spaceskip=\fontdimen2\font plus
\BIBentryALTinterwordstretchfactor\fontdimen3\font minus \fontdimen4\font\relax}
\providecommand{\BIBforeignlanguage}[2]{{%
\expandafter\ifx\csname l@#1\endcsname\relax
\typeout{** WARNING: IEEEtran.bst: No hyphenation pattern has been}%
\typeout{** loaded for the language `#1'. Using the pattern for}%
\typeout{** the default language instead.}%
\else
\language=\csname l@#1\endcsname
\fi
#2}}
\providecommand{\BIBdecl}{\relax}
\BIBdecl

\bibitem{Wang_2017_HDGE}
H.~Wang and Z.~Li, ``Region {{Representation Learning}} via {{Mobility Flow}},'' in \emph{Proceedings of the 2017 {{ACM}} on {{Conference}} on {{Information}} and {{Knowledge Management}}}.\hskip 1em plus 0.5em minus 0.4em\relax {Singapore Singapore}: {ACM}, Nov. 2017, pp. 237--246.

\bibitem{Zhang_2020_MVGRE}
M.~Zhang, T.~Li, Y.~Li, and P.~Hui, ``Multi-{{View Joint Graph Representation Learning}} for {{Urban Region Embedding}},'' in \emph{Proceedings of the {{Twenty-Ninth International Joint Conference}} on {{Artificial Intelligence}}}.\hskip 1em plus 0.5em minus 0.4em\relax {Yokohama, Japan}: {International Joint Conferences on Artificial Intelligence Organization}, Jul. 2020, pp. 4431--4437.

\bibitem{Jean_2018_Tile2Vec}
N.~Jean, S.~Wang, A.~Samar, G.~Azzari, D.~Lobell, and S.~Ermon, ``{{Tile2Vec}}: {{Unsupervised}} representation learning for spatially distributed data,'' May 2018.

\bibitem{Yao_2018_ZeMob}
Z.~Yao, Y.~Fu, B.~Liu, W.~Hu, and H.~Xiong, ``Representing {{Urban Functions}} through {{Zone Embedding}} with {{Human Mobility Patterns}},'' in \emph{Proceedings of the {{Twenty-Seventh International Joint Conference}} on {{Artificial Intelligence}}}.\hskip 1em plus 0.5em minus 0.4em\relax {Stockholm, Sweden}: {International Joint Conferences on Artificial Intelligence Organization}, Jul. 2018, pp. 3919--3925.

\bibitem{GPT}
\BIBentryALTinterwordspacing
A.~Radford and K.~Narasimhan, ``Improving language understanding by generative pre-training,'' 2018. [Online]. Available: \url{https://api.semanticscholar.org/CorpusID:49313245}
\BIBentrySTDinterwordspacing

\bibitem{GPT2}
\BIBentryALTinterwordspacing
A.~Radford, J.~Wu, R.~Child, D.~Luan, D.~Amodei, and I.~Sutskever, ``Language models are unsupervised multitask learners,'' 2019. [Online]. Available: \url{https://api.semanticscholar.org/CorpusID:160025533}
\BIBentrySTDinterwordspacing

\bibitem{MoCO}
K.~He, H.~Fan, Y.~Wu, S.~Xie, and R.~Girshick, ``Momentum contrast for unsupervised visual representation learning,'' in \emph{2020 IEEE/CVF Conference on Computer Vision and Pattern Recognition (CVPR)}, 2020, pp. 9726--9735.

\bibitem{SimCLR}
T.~Chen, S.~Kornblith, M.~Norouzi, and G.~Hinton, ``A simple framework for contrastive learning of visual representations,'' in \emph{Proceedings of the 37th International Conference on Machine Learning}, ser. ICML'20.\hskip 1em plus 0.5em minus 0.4em\relax JMLR.org, 2020.

\bibitem{TKDE_SSL_sur}
X.~Liu, F.~Zhang, Z.~Hou, L.~Mian, Z.~Wang, J.~Zhang, and J.~Tang, ``Self-supervised learning: Generative or contrastive,'' \emph{IEEE Transactions on Knowledge and Data Engineering}, vol.~35, no.~1, pp. 857--876, 2023.

\bibitem{Fu_2019_MVPN}
Y.~Fu, P.~Wang, J.~Du, L.~Wu, and X.~Li, ``Efficient {{Region Embedding}} with {{Multi-View Spatial Networks}}: {{A Perspective}} of {{Locality-Constrained Spatial Autocorrelations}},'' \emph{Proceedings of the AAAI Conference on Artificial Intelligence}, vol.~33, no.~01, pp. 906--913, Jul. 2019.

\bibitem{CGAL}
Y.~Zhang, Y.~Fu, P.~Wang, X.~Li, and Y.~Zheng, ``Unifying {{Inter-region Autocorrelation}} and {{Intra-region Structures}} for {{Spatial Embedding}} via {{Collective Adversarial Learning}},'' in \emph{Proceedings of the 25th {{ACM SIGKDD International Conference}} on {{Knowledge Discovery}} \& {{Data Mining}}}.\hskip 1em plus 0.5em minus 0.4em\relax {Anchorage AK USA}: {ACM}, Jul. 2019, pp. 1700--1708.

\bibitem{Wu_2022_MGFN}
S.~Wu, X.~Yan, X.~Fan, S.~Pan, S.~Zhu, C.~Zheng, M.~Cheng, and C.~Wang, ``Multi-{{Graph Fusion Networks}} for {{Urban Region Embedding}},'' May 2022.

\bibitem{ROMER}
\BIBentryALTinterwordspacing
W.~Chan and Q.~Ren, ``Region-wise attentive multi-view representation learning for urban region embedding,'' in \emph{Proceedings of the 32nd ACM International Conference on Information and Knowledge Management}, ser. CIKM '23.\hskip 1em plus 0.5em minus 0.4em\relax New York, NY, USA: Association for Computing Machinery, 2023, p. 3763–3767. [Online]. Available: \url{https://doi.org/10.1145/3583780.3615194}
\BIBentrySTDinterwordspacing

\bibitem{HREP}
S.~Zhou, D.~He, L.~Chen, S.~Shang, and P.~Han, ``Heterogeneous {{Region Embedding}} with {{Prompt Learning}},'' \emph{AAAI}, vol.~37, no.~4, pp. 4981--4989, Jun. 2023.

\bibitem{advaugmenre}
L.~Zhao, T.~Liu, X.~Peng, and D.~Metaxas, ``Maximum-entropy adversarial data augmentation for improved generalization and robustness,'' in \emph{Proceedings of the 34th International Conference on Neural Information Processing Systems}, ser. NIPS '20.\hskip 1em plus 0.5em minus 0.4em\relax Red Hook, NY, USA: Curran Associates Inc., 2020.

\bibitem{robustCL}
\BIBentryALTinterwordspacing
Z.~Jiang, T.~Chen, T.~Chen, and Z.~Wang, ``Robust pre-training by adversarial contrastive learning,'' in \emph{Advances in Neural Information Processing Systems}, H.~Larochelle, M.~Ranzato, R.~Hadsell, M.~Balcan, and H.~Lin, Eds., vol.~33.\hskip 1em plus 0.5em minus 0.4em\relax Curran Associates, Inc., 2020, pp. 16\,199--16\,210. [Online]. Available: \url{https://proceedings.neurips.cc/paper_files/paper/2020/file/ba7e36c43aff315c00ec2b8625e3b719-Paper.pdf}
\BIBentrySTDinterwordspacing

\bibitem{Kipf_2016_Variational}
T.~N. Kipf and M.~Welling, ``Variational {{Graph Auto-Encoders}},'' Nov. 2016.

\bibitem{Kipf_2017_SemiSupervised}
------, ``Semi-{{Supervised Classification}} with {{Graph Convolutional Networks}},'' Feb. 2017.

\bibitem{R-GCNs}
\BIBentryALTinterwordspacing
M.~Schlichtkrull, T.~N. Kipf, P.~Bloem, R.~van\&nbsp;den Berg, I.~Titov, and M.~Welling, ``Modeling relational data with graph convolutional networks,'' in \emph{The Semantic Web: 15th International Conference, ESWC 2018, Heraklion, Crete, Greece, June 3–7, 2018, Proceedings}.\hskip 1em plus 0.5em minus 0.4em\relax Berlin, Heidelberg: Springer-Verlag, 2018, p. 593–607. [Online]. Available: \url{https://doi.org/10.1007/978-3-319-93417-4_38}
\BIBentrySTDinterwordspacing

\bibitem{HetGNN}
\BIBentryALTinterwordspacing
C.~Zhang, D.~Song, C.~Huang, A.~Swami, and N.~V. Chawla, ``Heterogeneous graph neural network,'' in \emph{Proceedings of the 25th ACM SIGKDD International Conference on Knowledge Discovery \& Data Mining}, ser. KDD '19.\hskip 1em plus 0.5em minus 0.4em\relax New York, NY, USA: Association for Computing Machinery, 2019, p. 793–803. [Online]. Available: \url{https://doi.org/10.1145/3292500.3330961}
\BIBentrySTDinterwordspacing

\bibitem{ADGSTUDY}
\BIBentryALTinterwordspacing
W.~Jin, Y.~Li, H.~Xu, Y.~Wang, S.~Ji, C.~Aggarwal, and J.~Tang, ``Adversarial attacks and defenses on graphs,'' \emph{SIGKDD Explor. Newsl.}, vol.~22, no.~2, p. 19–34, Jan. 2021. [Online]. Available: \url{https://doi.org/10.1145/3447556.3447566}
\BIBentrySTDinterwordspacing

\bibitem{ADGCL}
S.~Suresh, P.~Li, C.~Hao, and J.~Neville, ``Adversarial {{Graph Augmentation}} to {{Improve Graph Contrastive Learning}},'' Nov. 2021.

\bibitem{RGCN}
\BIBentryALTinterwordspacing
D.~Zhu, Z.~Zhang, P.~Cui, and W.~Zhu, ``Robust graph convolutional networks against adversarial attacks,'' in \emph{Proceedings of the 25th ACM SIGKDD International Conference on Knowledge Discovery \& Data Mining}, ser. KDD '19.\hskip 1em plus 0.5em minus 0.4em\relax New York, NY, USA: Association for Computing Machinery, 2019, p. 1399–1407. [Online]. Available: \url{https://doi.org/10.1145/3292500.3330851}
\BIBentrySTDinterwordspacing

\bibitem{Hamilton_2018_GraphSAGE}
W.~L. Hamilton, R.~Ying, and J.~Leskovec, ``Inductive {{Representation Learning}} on {{Large Graphs}},'' Sep. 2018.

\bibitem{Velickovic_2018_GAT}
P.~Veli{\v c}kovi{\'c}, G.~Cucurull, A.~Casanova, A.~Romero, P.~Li{\`o}, and Y.~Bengio, ``Graph {{Attention Networks}},'' Feb. 2018.

\bibitem{CILP}
A.~Radford, J.~W. Kim, C.~Hallacy, A.~Ramesh, G.~Goh, S.~Agarwal, G.~Sastry, A.~Askell, P.~Mishkin, J.~Clark, G.~Krueger, and I.~Sutskever, ``Learning {{Transferable Visual Models From Natural Language Supervision}},'' Feb. 2021.

\bibitem{BYOL}
J.-B. Grill, F.~Strub, F.~Altché, C.~Tallec, P.~H. Richemond, E.~Buchatskaya, C.~Doersch, B.~A. Pires, Z.~D. Guo, M.~G. Azar, B.~Piot, K.~Kavukcuoglu, R.~Munos, and M.~Valko, ``Bootstrap your own latent: A new approach to self-supervised learning,'' 2020.

\bibitem{Xu__InfoGCL}
\BIBentryALTinterwordspacing
D.~Xu, W.~Cheng, D.~Luo, H.~Chen, and X.~Zhang, ``Infogcl: Information-aware graph contrastive learning,'' 2021. [Online]. Available: \url{https://arxiv.org/abs/2110.15438}
\BIBentrySTDinterwordspacing

\bibitem{GraphCL}
\BIBentryALTinterwordspacing
Y.~You, T.~Chen, Y.~Sui, T.~Chen, Z.~Wang, and Y.~Shen, ``Graph contrastive learning with augmentations,'' 2021. [Online]. Available: \url{https://arxiv.org/abs/2010.13902}
\BIBentrySTDinterwordspacing

\bibitem{ResNet}
K.~He, X.~Zhang, S.~Ren, and J.~Sun, ``Deep residual learning for image recognition,'' in \emph{2016 IEEE Conference on Computer Vision and Pattern Recognition (CVPR)}, 2016, pp. 770--778.

\bibitem{TPAMICLSV}
Y.~Xie, Z.~Xu, J.~Zhang, Z.~Wang, and S.~Ji, ``Self-supervised learning of graph neural networks: A unified review,'' \emph{IEEE Transactions on Pattern Analysis and Machine Intelligence}, vol.~45, no.~2, pp. 2412--2429, 2023.

\bibitem{oord2019representation}
A.~van~den Oord, Y.~Li, and O.~Vinyals, ``Representation learning with contrastive predictive coding,'' 2019.

\bibitem{Berg__New}
B.~F. Berg, ``New {{York City Politics}} : {{Governing Gotham}},'' \emph{Rutgers University Press}, p. 352, 2007.

\bibitem{Tang_2015_LINE}
\BIBentryALTinterwordspacing
J.~Tang, M.~Qu, M.~Wang, M.~Zhang, J.~Yan, and Q.~Mei, ``Line: Large-scale information network embedding,'' in \emph{Proceedings of the 24th International Conference on World Wide Web}, ser. WWW '15.\hskip 1em plus 0.5em minus 0.4em\relax Republic and Canton of Geneva, CHE: International World Wide Web Conferences Steering Committee, 2015, p. 1067–1077. [Online]. Available: \url{https://doi.org/10.1145/2736277.2741093}
\BIBentrySTDinterwordspacing

\bibitem{Grover_2016_node2vec}
A.~Grover and J.~Leskovec, ``Node2vec: {{Scalable Feature Learning}} for {{Networks}},'' Jul. 2016.

\bibitem{GraphSAGE}
\BIBentryALTinterwordspacing
W.~Hamilton, Z.~Ying, and J.~Leskovec, ``Inductive representation learning on large graphs,'' in \emph{Advances in Neural Information Processing Systems}, I.~Guyon, U.~V. Luxburg, S.~Bengio, H.~Wallach, R.~Fergus, S.~Vishwanathan, and R.~Garnett, Eds., vol.~30.\hskip 1em plus 0.5em minus 0.4em\relax Curran Associates, Inc., 2017. [Online]. Available: \url{https://proceedings.neurips.cc/paper_files/paper/2017/file/5dd9db5e033da9c6fb5ba83c7a7ebea9-Paper.pdf}
\BIBentrySTDinterwordspacing

\bibitem{goodfellow2015explaining}
I.~J. Goodfellow, J.~Shlens, and C.~Szegedy, ``Explaining and harnessing adversarial examples,'' in \emph{Proceedings of the International Conference on Machine Learning (ICML)}, 2015.

\bibitem{madry2018towards}
A.~Madry, M.~Makelov, L.~Schmidt, D.~Tsipras, and A.~Vladu, ``Towards deep learning models resistant to adversarial attacks,'' in \emph{Proceedings of the International Conference on Machine Learning (ICML)}, 2018.

\bibitem{pasionnoise}
A.~Papoulis and S.~Pillai, \emph{Probability, Random Variables, and Stochastic Processes}, ser. {{McGraw-Hill}} Series in Electrical and Computer Engineering.\hskip 1em plus 0.5em minus 0.4em\relax Online: McGraw-Hill, 2002.

\end{thebibliography}
\bibliographystyle{IEEEtran}

\newpage

 





\end{document}